\documentclass[10pt,twocolumn,letterpaper]{article}

\usepackage[pagenumbers]{cvpr} 
\usepackage{float}
\usepackage{algorithm,algorithmicx,algpseudocode}
\usepackage{bm}
\usepackage{setspace}
\usepackage{amsmath}  
\usepackage{amssymb}   
\usepackage{amsthm} 
\newtheorem{thm}{Theorem}
\newtheorem{thm2}{Theorem}
\newtheorem{Proposition}[thm]{Proposition}
\newtheorem{Corollary}[thm2]{Corollary}
\newtheorem{Proposition_proof}{Proposition}
\newtheorem{Lemma}{Lemma}

\usepackage[dvipsnames]{xcolor}

\definecolor{cvprblue}{rgb}{0.21,0.49,0.74}
\usepackage[pagebackref,breaklinks,colorlinks,citecolor=cvprblue]{hyperref}
\usepackage{pgfplots}
\usepackage{multirow}
\usepackage{makecell}
\def\paperID{3588} 
\def\confName{CVPR}
\def\confYear{2024}

\title{Tackling the Singularities at the Endpoints of Time Intervals in Diffusion Models}

\author{Pengze Zhang\textsuperscript{1}$\footnotemark[1]$\ \ \ \ \ \ Hubery Yin\textsuperscript{2}$\footnotemark[1]$\ \ \ \ \ \ Chen Li\textsuperscript{2}\ \ \ \ \ \ Xiaohua Xie\textsuperscript{1}$\footnotemark[2]$\\
\textsuperscript{1}Sun Yat-sen University \ \ \ \ \ \ \textsuperscript{2}Wechat, Tencent Inc.\\
{\tt\small zhangpz3@mail2.sysu.edu.cn, \{hubery, chaselli\}@tencent.com, xiexiaoh6@mail.sysu.edu.cn} \\
\tt\small\url{https://pangzecheung.github.io/SingDiffusion/}
}

\usepackage[normalem]{ulem}

\begin{document}

\maketitle
\renewcommand{\thefootnote}{\fnsymbol{footnote}}  
\footnotetext[1]{Equal contribution. This work was done when Pengze Zhang was an intern at WeChat.}
\footnotetext[2]{Corresponding Author.}
\renewcommand{\thefootnote}{\arabic{footnote}} 

\begin{abstract}
Most diffusion models assume that the reverse process adheres to a Gaussian distribution. However, this approximation has not been rigorously validated, especially at singularities, where $t=0$ and $t=1$. 
Improperly dealing with such singularities leads to an average brightness issue in applications, and limits the generation of images with extreme brightness or darkness.
We primarily focus on tackling singularities from both theoretical and practical perspectives.
Initially, we establish the error bounds for the reverse process approximation, and showcase its Gaussian characteristics at singularity time steps. 
Based on this theoretical insight, we confirm the singularity at $t=1$ is conditionally removable while it at $t=0$ is an inherent property. 
Upon these significant conclusions, we propose a novel plug-and-play method \textbf{SingDiffusion} to address the initial singular time step sampling, which not only effectively resolves the average brightness issue for a wide range of diffusion models without extra training efforts, but also enhances their generation capability in achieving notable lower FID scores.
\end{abstract}    
\section{Introduction}
\label{sec:intro}

Diffusion models, generating samples from initial noise by learning a reverse  diffusion process, have achieved remarkable success in multi-modality content generation, such as image generation \cite{dhariwal2021diffusion, nichol2021glide, pmlr-v139-ramesh21a,saharia2022photorealistic,campbell2022a,zhang2023formulating}, audio generation \cite{nichol2021glide, kong2020diffwave, popov2021grad}, and video generation \cite{ho2022video}. 
These achievements owe much to several fundamental theoretical research, namely Denoising Diffusion Probabilistic Modeling (DDPM) \cite{sohl2015deep, ho2020denoising}, Stochastic Differential Equations (SDE) \cite{song2020score}, and Ordinary Differential Equations (ODE) \cite{song2020score, song2020denoising}. 
These approaches are all based on the assumption that the reverse diffusion process shares the same functional form as the forward process.  
Although an indirect validation of this assumption is given by Song et al. \cite{song2020score}, it heavily relies on the existence of solutions to the Kolmogorov forward and backward equations \cite{anderson1982reverse}, which encounters singularities at the endpoints of time intervals where $t=0$ and $t=1$.

The singularity issue is not only a gap in the theoretical formulation of current diffusion models, but also affects the quality of the content they generate.
Current applications simply ignore singularity points in their implementation \cite{von-platen-etal-2022-diffusers, Rombach_2022_CVPR, podell2023sdxl} and restrict the time interval to \ $[\varepsilon_1, 1-\varepsilon_2]$. As a result, the average brightness of the generated images typically hovers around 0 \cite{nicholas2023diffusion, lin2023common} (normalizing brightness to $[-1, 1]$).
For example, as shown in Fig. \ref{fig:fig1}, existing pre-trained diffusion models, such as 
\emph{Stable Diffusion 1.5} (SD-1.5) \cite{Rombach_2022_CVPR} and \emph{Stable Diffusion 2.0-base} (SD-2.0-base), fail in generating images with pure white or black backgrounds.
To address this challenge, Guttenberg et al. \cite{nicholas2023diffusion} add extra offset noise during the training process to allow the network could learn the overall brightness changes of the image. 
Unfortunately, the offset noise usually disrupts the pre-defined marginal probability distribution and further invalidates the original sampling formula.
Lin et al. \cite{lin2023common} re-scales the noise schedule to enforce a zero terminal signal-to-noise ratio, and employ the $v$-prediction technique \cite{salimans2022progressive} to circumvent the issue of division by zero at $t=1$. 
However, this method only supports models in $v$-prediction manner and requires substantial training to fine-tune the entire model.
Therefore, it is advantageous to devise a plug-and-play method which effectively deals with the singularity issue for any practicable diffusion model without extra training efforts.

In this paper, we begin with a theoretical exploration of the singularity issue at the endpoints of time intervals in diffusion model, and then devise a novel plug-and-play module to address the accompanying average brightness problem in image generation. Through establishing mathematical error bounds rigorously, we first prove the approximate Gaussian characteristics of the reverse diffusion process at all sampling time steps, especially where the singularity issue appears.
Following this, we conduct a thorough analysis of this approximation in the vicinity of singularities, and arrive at two significant conclusions: 1) by computing the limit of `zero divided by zero' at $t=1$, we confirm the corresponding initial singularity is removable; 2) the singularity at $t=0$ is an inherent property of diffusion models, thus we should adhere to this form rather than simply avoiding the singularity.
Following the aforementioned theoretical analysis, we propose the SingDiffusion method, specifically tailored to address the challenge of sampling during the initial singularity. 
Especially, this method can seamlessly integrate into existing sampling processes in a plug-and-play manner without requiring additional training efforts.

As demonstrated in Fig. \ref{fig:fig1} and experiments in the appendix, our novel plug-and-play initial sampling step can effectively resolve the average brightness issue. It can be easily applied to a wide range of diffusion models, including the SD-1.5 and SD-2.0-base with $\epsilon$-prediction, SD-2.0 with $v$-prediction, and various pre-trained models available on the CIVITAI website$\footnote{https://civitai.com/}$, thanks to its one-time training strategy. Furthermore, our method can generally enhance the generative capabilities of these pre-trained diffusion models in notably improving the FID \cite{heusel2017gans} scores at the same CLIP \cite{radford2021learning} level on the COCO dataset \cite{lin2014microsoft}.


\section{Related Work}

\subsection{Reverse process approximation}
Denoising Diffusion Probability Models (DDPM) \cite{sohl2015deep, ho2020denoising} establish a hand-designed forward Markov chain in discrete time, and model the distribution of data in the reverse process.
In contrast, Song et al. \cite{song2020score} establish the diffusion model in continuous time, framed as a Stochastic Differential Equation (SDE). 
Moreover, they reveal an Ordinary Differential Equation (ODE) termed `probability flow', sharing the same single-time marginal distribution as the original SDE.  
Notably, ODE also has discrete counterparts known as Denoising Diffusion Implicit Models (DDIM) \cite{song2020denoising}.

The assumption that the reversal of the diffusion process has an identical functional form to the forward process is fundamental in the aforementioned methods.
Several studies aim to prove this assumption. 
Song et al. \cite{song2020score} indirectly substantiate this in continuous time by introducing a reverse SDE. 
Nevertheless, they do not provide error bounds in discrete-time cases, leaving this assumption unverified at discrete-time steps. 
Additionally, the treatment of singularities at $t=0$ and $t=1$ is not addressed in their work.
McAllester et al. \cite{mcallester2023mathematics} provide proof in discrete time, showing the density of the reverse process as a mixture of Gaussian distributions.
Despite this, it doesn't qualify as a pure Gaussian distribution.
In contrast, to fill this theoretical gap, our approach directly substantiates this assumption by establishing error bounds for the reverse process approximation at both non-singular and singular time steps.

\subsection{Singularities in diffusion model}
Several studies have focused on investigating the singularity occurring at $t=0$. 
Song et al. \cite{song2020score} attempt to bypass this singularity by initiating their analysis at $t=\varepsilon > 0$ instead of $t=0$. Therefore, this approach didn't effectively address the core singularity problem.
Dockhorn et al. \cite{dockhorn2021score} propose a diffusion model on the product space of position and velocity, and avoid the singularity through hybrid score matching. Nevertheless, this approach lacks compatibility with DDPM, SDE and ODE due to the incorporation of the velocity space.
Lu et al. \cite{lu2023mathematical} employ the $x$-prediction method to mitigate singularity during training, but did not address the singularity issue during the sampling process.

Therefore, a comprehensive solution to the singularity at $t=0$ is still pending. 
Moreover, the singularity at $t=1$ remains unexplored. 
To tackle these issues, we thoroughly investigate the sampling process at both $t=0$ and $t=1$, and offer theoretical solutions.

\subsection{Average brightness issue}

Diffusion models have shown significant comprehensive quality and controllability in computer vision, including text-to-image generation \cite{Rombach_2022_CVPR,pmlr-v139-ramesh21a,saharia2022photorealistic,dhariwal2021diffusion, nichol2021glide, ho2021classifierfree}, image editing \cite{hertz2023prompttoprompt, ruiz2023dreambooth, kawar2023imagic, avrahami2022blended, meng2021sdedit,Yang_2023_CVPR}, image-to-image translation \cite{saharia2022palette, su2022dual, Zhang_2023_ICCV}, surpassing previous generative models \cite{NIPS2014_5ca3e9b1,Zhang_2022_CVPR,Esser_2021_CVPR,taming}.
However, most existing diffusion models ignore the sampling at the initial singular time step, resulting in the inability to generate bright and dark images, i.e., the average brightness issue. 
Adding offset noise \cite{nicholas2023diffusion} and employing $v$-prediction \cite{lin2023common, salimans2022progressive} are two ways to tackle this problem.
However, these methods require fine-tuning for each existing model, consuming a substantial amount of time and limiting their applicability. 
In contrast, we propose a novel plug-and-play solution targeting the core of the average brightness issue, i.e., the singularity at the initial time step.
Our method not only empowers the majority of existing pre-trained models to effectively generate images with the desired brightness level, but also significantly enhances their generative capabilities.
\section{Method}
To facilitate the clarity of our exploration into singularities from theoretical and practical angles, this section is organized as follows: 1) We start by introducing background and symbols in the Preliminaries. 2) Next, we derive error bounds for the reverse process approximation, confirming its Gaussian characteristics at both regular and singular time steps. 3) We then theoretically analyze and handle the sampling at singular time steps, i.e., $t=0$ and $t=1$. 4) Lastly, based on our previous analysis, we propose a plug-and-play method to address initial singularity time step sampling, effectively resolving the average brightness issue.

\subsection{Preliminaries}\label{sec:Preliminaries}

In the realm of generative models, we are consistently provided with a set of training samples denoted as $\{ y_i \in \mathbb{R}^d \}_{i=1}^N$,
which are inherently characterized by a distribution given by \cite{karras2022elucidating}: 
\begin{equation}\label{init_dist}
    p(x, t=0) = \frac{1}{N} \sum_{i=1}^N \delta (x - y_i),
\end{equation}
where $\delta(x)$ denotes the Dirac delta function. 

Consider a continuous-time Gaussian diffusion process within the interval $0\leq t \leq 1$. Following \cite{NEURIPS2021_b578f2a5}, the distribution of $x_t$ conditioned on $x_0$ is written by $p(x_t|x_0) = \mathcal{N}(\alpha_{t}x_0, \sigma_t^{2}I)$, where $\alpha_t$ and $\sigma_t$ are positive scalar functions of $t$ satisfying $\alpha_t^2 + \sigma_t^2 = 1$, and $\alpha_t$ decreases monotonically from 1 to 0 over time $t$; 
the distribution of $x_t$ conditioned on $x_s$ is represented as $p(x_t|x_s) = \mathcal{N}(\alpha_{t|s}x_s, \sigma_{t|s}^{2}I)$, where $0 \leq s < t \leq 1$, $\alpha_{t|s} = \alpha_t / \alpha_s$, $\sigma_{t|s}^{2} = 1 - \alpha_{t|s}^{2}$.
Consequently, the forward process is derived as follows:
\begin{equation}\label{ou_sol}
    x_t = \alpha_{t|s} x_s + \sqrt{1 - \alpha_{t|s}^{2}} z_s,
\end{equation}
where the set $\{z_t\}_{t=0}^1$ comprises independent standard Gaussian random variables. 

For a discrete-time diffusion process with $T$ steps, the time $i \in \{0,1,...T\}$ corresponds to $t$ in the continuous case as $i = t \times T$.
Defining $\hat{\beta}_i = 1 - \hat{\alpha}_{i|i-1}^2$, where $\hat{\alpha}_{i|i-1} = \alpha_{i / T|(i-1)/T}$, and ` $\hat{}$ ' denotes the symbol in the discrete-time process, the forward process in Eq. \ref{ou_sol} can be rewritten as:
\begin{equation}
    \hat{x}_i = \sqrt{1 - \hat{\beta}_i} \hat{x}_{i-1} + \sqrt{\hat{\beta}_i} \hat{z}_{i-1}, 
\end{equation}
which is equivalent to the forward process outlined in \cite{ho2020denoising}.

Taking into account the initial distribution (Eq. \ref{init_dist}), the single-time marginal distribution of $x_t$ is given by:
\begin{equation}\label{stm}
    p(x_t, t) = \frac{1}{N} \sum_i (2 \pi \sigma_t^2)^{-\frac{d}{2}} \exp(-\frac{(x_t - \alpha_t y_i)^2}{2 \sigma_t^2}),
\end{equation}
where $d$ is the dimension of the training samples. As a result, the reverse process can be derived using Bayes' rule:
\begin{equation}\label{bayes}
\begin{aligned}
    &p(x_s | x_t) = p(x_t| x_s) \frac{p(x_s, s)}{p(x_t, t)} =(2\pi \sigma_{s|t})^{-\frac{d}{2}} \\
    &~~\sum_i \exp(-\frac{1}{2\sigma_{s|t}^2} (x_s - \frac{\alpha_{t|s} \sigma_s^2 x_t}{\sigma_t^2} - \frac{\alpha_s \sigma_{t|s}^2 y_i}{\sigma_t^2})^2) w_i(x_t, t),
\end{aligned}
\end{equation}
where $\sigma_{s|t}^2 = \sigma_{t|s}^2 \frac{\sigma_s^2}{\sigma_t^2}$, and $w_i(x_t, t) = \frac{\exp(-\frac{(x_t - \alpha_t y_i)^2}{2 \sigma_t^2})}{\sum_j \exp(-\frac{(x_t - \alpha_t y_j)^2}{2 \sigma_t^2})}$.
\subsection{Error bound estimation} \label{sec:error}
Existing diffusion models, such as \cite{sohl2015deep,ho2020denoising} are based on an assumption that the reverse process in Eq. \ref{bayes} can be approximated by a Gaussian distribution, when $\hat{\beta}_i$ is small, as given by:
\begin{equation}\label{dis_inv} 
\begin{aligned}
    \tilde{p}(x_s| x_t) &= (2\pi \sigma_{s|t}^2)^{-\frac{d}{2}} \\
    &\exp(-\frac{1}{2\sigma_{s|t}^2} (x_s - \frac{\alpha_{t|s} \sigma_s^2 x_t}{\sigma_t^2} -\frac{\alpha_s \sigma_{t|s}^2 \bar{y}(x_t, t)}{\sigma_t^2})^2),
\end{aligned}
\end{equation}
where $\bar{y}(x_t, t) = \sum_i w_i(x_t, t) y_i$.\label{y_bar} However, these studies did not furnish the error bounds to support this assumption.
To address this theoretical gap, we estimate the error bound as follows.
\begin{Proposition}[\textbf{Error Bound Estimated by $\sigma_{s|t}$}]
\label{prop1}$\forall s \in (0, 1)$, $\exists \tau \in (s, 1)$ and $C > 0$, such that $\forall t \in (s, \tau]$,
$\int_{\mathbb{R}^{d}} |p(x_s| x_t) - \tilde{p}(x_s| x_t)| d x_s < C \sqrt{\sigma_{s|t}}$.
\end{Proposition}
Proposition \ref{prop1} demonstrates that when $\sigma_{s|t}$ is small, the forward and reverse processes share the same form, i.e., Gaussian distribution. 
Since $\hat{\beta}_i = \frac{\sigma_{i / T}}{\sigma_{(i-1)/T}}\sigma_{(i-1)/T | i / T}$ features a term in the $\sigma_{s|t}$ form, also supports the inference that this assumption remains valid when $\hat{\beta}_i$ is small, as highlighted in \cite{sohl2015deep,ho2020denoising}.
It is worth noting that this error is bounded by $\sigma_{s|t}$ instead of $\hat{\beta}_i$ strictly.

However, the error bound estimated by $\sigma_{s|t}$ in Proposition \ref{prop1} is not sufficient to prove the assumption at the singularity time step $t=1$. 
The reason is that when $t=1$, $\sigma_{s|t}$ approaches 1 as $s \to 1$, which is not a small value. 
Consequently, the error bound at $t=1$ in Proposition \ref{prop1} remains non-negligible.

To tackle this issue, we instead utilize $\alpha_s$ to bound the error at time step $t=1$, and present a new proposition:
\begin{Proposition} [\textbf{Error Bound Estimated by $\alpha_s$}] \label{prop2}$\exists \nu \in (0, 1)$ and $C > 0$, such that $\forall \nu \leq s < t \leq 1$,
$\int_{\mathbb{R}^{d}} |p(x_s| x_t) - \tilde{p}(x_s| x_t)| d x_s < C \sqrt{\alpha_s}$.
\end{Proposition}
According to Proposition \ref{prop2}, setting $t = 1$, it has $\alpha_s \to 0$ as $s \to 1$. As a result, the error bound at $t=1$ assures a small value, affirming the validation of the Gaussian approximation assumption at $t=1$.

In sum, through Proposition \ref{prop1} and \ref{prop2}, we have established that the reverse process of the diffusion model can be approximated by Gaussian distribution across all time steps.

\subsection{Tackling the singularities}\label{sec:sing}
With the theoretical foundation provided in Section \ref{sec:error}, we can delve into the analysis of singular time steps using Eq. \ref{dis_inv}. 
It is worth noting that this section will mainly focus on addressing the singularities present in the discrete-time diffusion model \cite{ho2020denoising}, while the treatment of the continuous case \cite{song2020score} will be deferred to the appendix.

\subsubsection{The singularities at $t=1$} \label{sing_t1}

Drawing from Eq. \ref{dis_inv}, a straightforward way is to train a neural network $\bar{y}_\theta(x_t, t)$ for estimating $\bar{y}(x_t, t)$, known as $x$-prediction. 
This approach ensures that the approximated reverse process avoids encountering a singularity at $t=1$. 
However, for stable training \cite{sohl2015deep}, the mainstream choice among current diffusion models is $\epsilon$-prediction.
It utilizes a neural network $\epsilon_\theta(x_t, t)$ for estimating $\epsilon(x_t, t) = \frac{x_t - \alpha_t \bar{y}(x_t, t)}{\sigma_t}$, which will encounter singularity.
More specifically, substituting $\epsilon(x_t, t)$ for $\bar{y}(x_t, t)$ yields the equation:
\begin{equation}
    x_s = \frac{\alpha_{t|s} \sigma_s^2}{\sigma_t^2} x_t + \frac{\alpha_s \sigma_{t|s}^2}{\sigma_t^2} \frac{x_t - \sigma_t \epsilon(x_t, t)}{\alpha_t} + \sigma_{s|t} z_t.
\end{equation}
At $t=1$, the denominator $\alpha_t$ becomes 0, resulting in a division-by-zero singularity.
Theoretically, this singularity is removable because the limit exists:
\begin{equation}
    \lim_{t \to 1^-} \frac{x_t - \sigma_t \epsilon(x_t, t)}{\alpha_t} = \lim_{t \to 1^-} \bar{y}(x_t,t) = \frac{1}{N} \sum_i y_i.
\end{equation}
Regrettably, computing this limit during inference is unfeasible, since $\epsilon(x_1, 1) = x_1$ loses all information of the correct sampling direction.
Conversely, $\bar{y}(x_t, t)$ retains the correct sampling direction for all $t \in [0, 1]$.
Particularly at $t=1$, $\bar{y}(x_1, 1) = \frac{1}{N} \sum_i y_i$ encapsulates the information for the initial inference step.
Therefore, leveraging $x$-prediction at the initial time step proves more advantageous than $\epsilon$-prediction.

\subsubsection{The singularities at $t=0$}
When $s=0$ and $t$ is small, the distribution in Eq. \ref{dis_inv} degenerates into a singular distribution, i.e., a Gaussian with zero variance, resembling a Dirac delta:
\begin{equation}
\begin{aligned}\label{dis_t1}
    \tilde{p}(x_0 | x_t) = \delta(x_0 - y_{j_0}),
\end{aligned}
\end{equation}
where $j_0 = \mathop{\arg\min}\limits_{j} |x_t - \alpha_t y_j|$.
This singularity directs the sampling process to converge at the correct point $y_{j_0} = \bar{y}(x_0, 0)$. 
Therefore, the singularity at $t=0$ is an inherent characteristic of diffusion models that do not require avoidance as long we use suitable sampling techniques. 
For instance, in the final step of the original DDPM sampling method, it has $x_0 = \bar{y}(x_t, t)$. 
When $t$ is small, $\bar{y}(x_t, t) \approx \bar{y}(x_0, 0)$, making this process equivalent to Eq. \ref{dis_t1}. Thus, there is no need to avoid the singularity.

Besides, we also arrived at this conclusion within the continuous diffusion model, i.e., SDE. 
More detailed elaboration on this topic can be found in the appendix.

\begin{figure}[!t]
\centering
\includegraphics[width=0.48\textwidth]{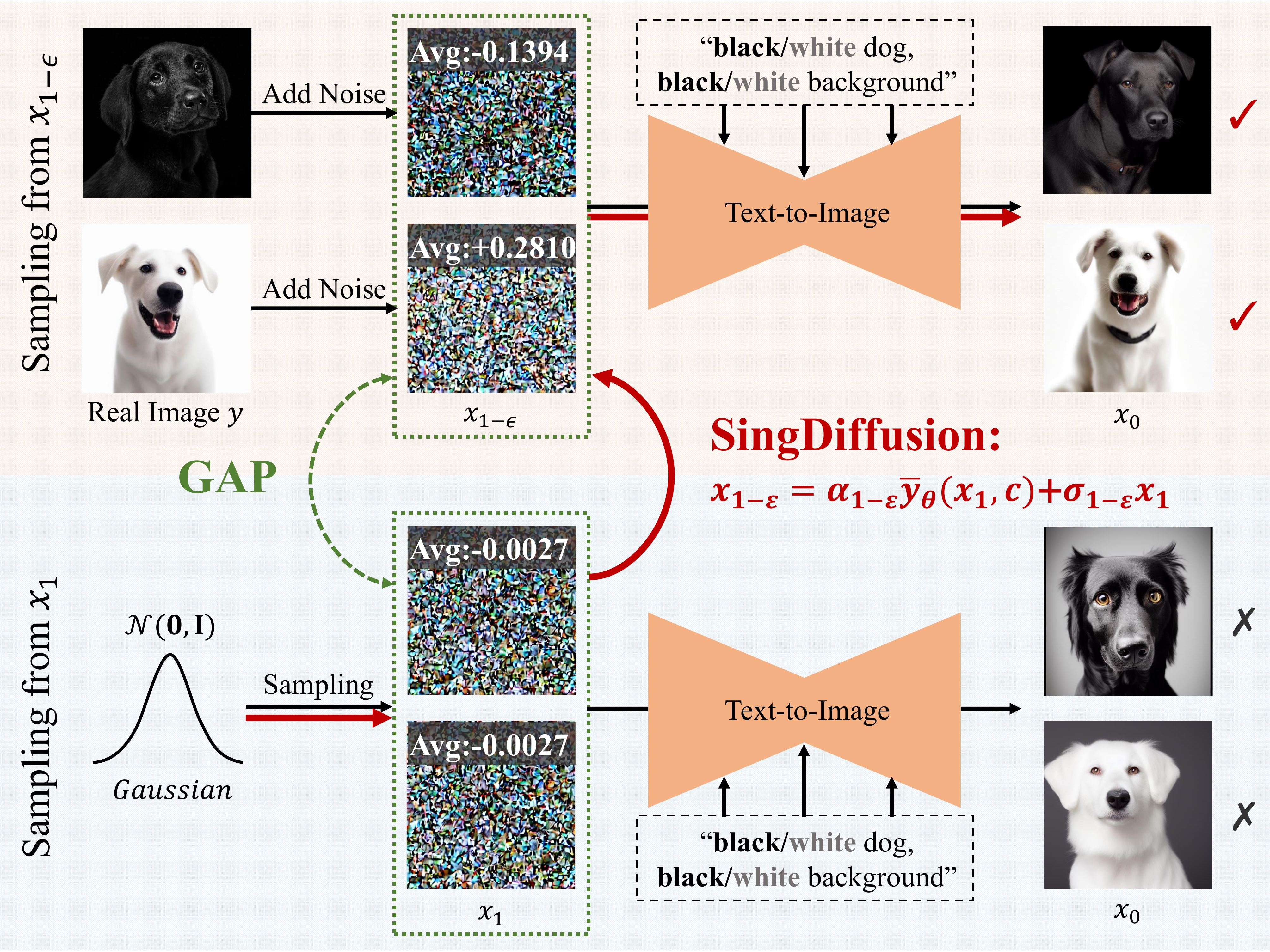} 
\caption{Illusion of the gap between the sampling from $x_{1-\varepsilon}$ and $x_{1}$. Due to the lack of consideration of singular time step sampling in most of the existing methods, they will encounter the average brightness issue. To tackle this, we propose a plug-and-play SingDiffusion method (highlighted in red) to bridge this gap.}
\label{fig:gap}
\end{figure}

\subsection{SingDiffusion}
In addition to theoretical issues, the singular sampling issue can also cause average brightness issue in applications, as depicted in Fig. \ref{fig:gap}.
This is mainly because most of the existing method sample images starting at time step $1-\varepsilon$ using a standard Gaussian distribution, which significantly diverges from the true distribution $p(x_{1-\varepsilon}, 1-\varepsilon)$.
To validate this, we select two images (a black dog against a black background and a white dog against a white background). 
Then we diffuse them for a time period of $1-\varepsilon$ respectively, and calculate the mean value of their latent code. 
It can be seen that the mean of these two latent codes are -0.14 and +0.28, which are significantly different from the samples obtained from the standard Gaussian distribution. 
Moreover, as evident from the visualization in Fig. \ref{fig:gap}, the latent code ($x_{1-\varepsilon}$) of the black dog and white dog are notably darker and brighter compared to $x_{1}$ from the Gaussian distribution respectively. 
Under such distributional differences, according to Proposition \ref{prop3}, employing a Gaussian distribution at $t=1-\epsilon$ is equivalent to generating images towards an average brightness of 0 at $t=1$ (with grayscale ranging from [-1, 1]). 
Consequently, current methods encounter challenges in generating dark or bright images.
\begin{Proposition} \label{prop3}Setting $x_{1-\epsilon} \sim \mathcal{N}(0, I)$ is equivalent to sampling the value from standard Gaussian as $\bar{y}(x_{1}, 1)$ at $t=1$.
\end{Proposition}

Based on our analysis of sampling at singular time steps in Section \ref{sing_t1}, we propose a novel plug-and-play method SingDiffusion to fill the gap at $t=1$, thus solving the problem of average brightness issue.
Considering a pre-trained model $\epsilon_\theta(x_t, t)$ afflicted by the singularity due to division by zero, we proceed to train a model using $x$-prediction at $t=1$.
The algorithm of our training and sampling process are shown in Algorithm \ref{alg:training} and Algorithm \ref{alg:sampling}. 
Firstly, for image-prompt data pairs ($x_0, c$) in the training process, we use a U-net \cite{Unet} $\bar{y}_\theta$ to fit $\bar{y}(x_1, t=1, c)$, where $\bar{y}(x_t, t, c)$ is the extension of $\bar{y}(x_t, t)$ defined in Section \ref{y_bar} under the condition $c$. 
The loss function can be written as:
\begin{equation}
    {\mathcal{L}={\mathbb{E}_{x_0, c \sim p(x_0, c), x_1 \sim \mathcal{N}(\bm{0},\bm{I})} ||\bar{y}_\theta (x_1, c) - x_0||}^2}.
    \label{eq:loss}
\end{equation}
As our model is only amortized with respect to text embeddings and not the time step, we omit the variable $t$ for $\bar{y}_\theta$, and the U-net does not necessitate a complex architecture.

In the sampling process, we use the new model $\bar{y}_\theta$ in the initial time-step with a DDIM scheduler:
\begin{equation}
    x_{1 - \varepsilon} = \alpha_{1-\varepsilon} \bar{y}_\theta(x_1, c) + \sigma_{1-\varepsilon} x_1.
\end{equation}
This equation guarantees $x_{1-\varepsilon}$ adheres to the distribution $p(x_{1 - \varepsilon}, 1-\varepsilon)$. 
Following this, we utilize the existing pre-trained model $\epsilon_\theta(x_t, t, c)$/$v_\theta(x_t, t, c)$ in $\epsilon$-prediction/$v$-prediction manner to perform the subsequent sampling steps until it generates $x_0$.
It's worth noting that our method is solely involved in the sampling at $t=1$, independent of the subsequent sampling process. 
Consequently, our approach can be once-trained and seamlessly integrated into the majority of diffusion models.
\begin{algorithm}[t]
\begin{algorithmic}[1]
\Repeat
\State $x_0,c \sim p(x_0, c)$, $x_1\sim\mathcal{N}(\bm{0},\bm{I})$
\State Take gradient descent step on $\nabla_\theta \left\| \bar{y}_\theta(x_1, c) - x_0 \right\|^2$

\Until{converged}
 \caption{Training process of SingDiffusion}
 \label{alg:training}
 \end{algorithmic}
\end{algorithm}

\begin{algorithm}[t]
\begin{algorithmic}[1]
\State $x_1\sim\mathcal{N}(\bm{0},\bm{I})$
\State $\varepsilon = 1/T$
\State $x_{1-\varepsilon} = \alpha_{1-\varepsilon}\bar{y}_\theta(x_1, c) + \sigma_{1-\varepsilon}x_1$
\For{$t=1- \varepsilon, \dotsc, \varepsilon$}
    \State Calculate $x_{t - \varepsilon}$ using existing sampling algorithms
\EndFor
\State \textbf{return} $x_0$
 \caption{Sampling process of SingDiffusion}
 \label{alg:sampling}
 \end{algorithmic}
\end{algorithm}

For further improving matching between generated images and input prompts, existing diffusion models typically incorporate classifier-free techniques \cite{nichol2021glide}:
\begin{equation}
    o_{guidance} = o_{neg} + w \times (o_{pos} - o_{neg}),
\end{equation}
where $w \geq 1$ represents the guidance scale, $o$ signifies the output of the diffusion model which can be either $\bar{y}_\theta$, $\epsilon_\theta$ or $v_\theta$, `pos' and `neg' refer to the outputs corresponding to positive and negative prompts respectively. 
Nevertheless, we notice that when applying this technique at the initial singular time step, the influence of the guidance scale $w$ predominates the results due to the greater directional difference between the negative and positive outputs. 
To tackle this challenge, we implement a straightforward yet highly effective normalization method:
\begin{equation}
{\bar{y}_{\theta_{guidance}} = [\bar{y}_{\theta_{neg}} + w \times (\bar{y}_{\theta_{pos}} - \bar{y}_{\theta_{neg}})] / w.}
\label{eq:guidance}
\end{equation}

\begin{figure*}[!t]
\centering
\includegraphics[width=1\textwidth]{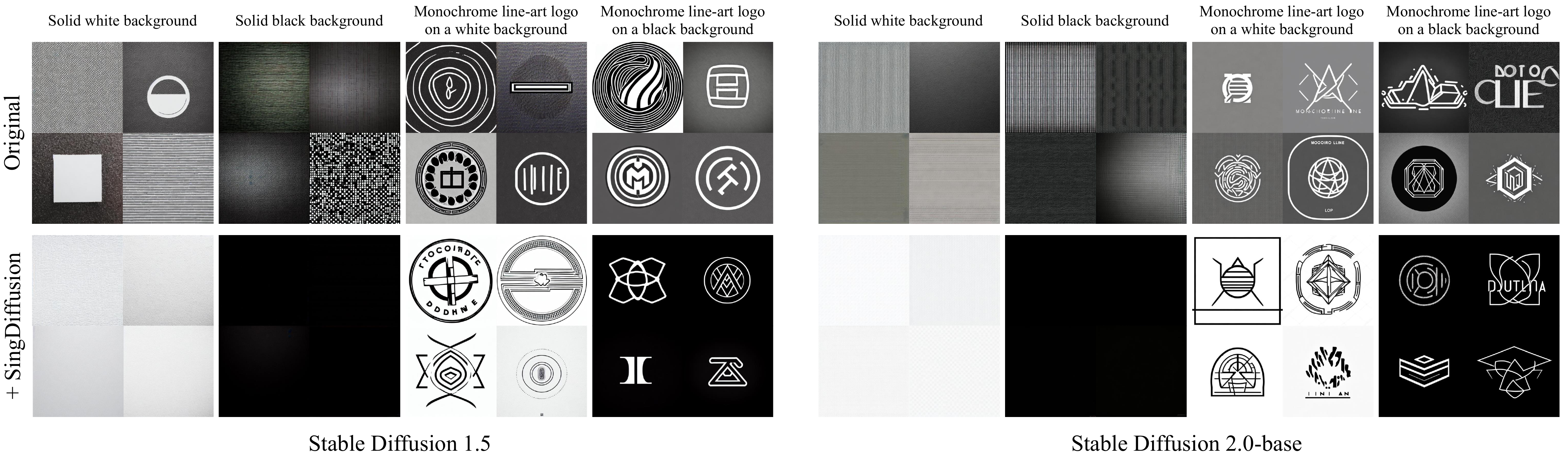} 
\caption{Comparison of stable diffusion models and SingDiffusion on average brightness issue.}
\label{fig:black}
\end{figure*}

\begin{figure*}[!t]
\centering
\center
\begin{tikzpicture}[font=\small,scale=0.8]
         \begin{axis}[
            ymin=9.3, ymax=17.5,
            ytick={10,...,14,15,16,17},
            xmin=28.7, xmax=31.4,
            ylabel=y, 
            ylabel=FID score $\downarrow$, y label style={at={(0.05,0.5)}},
               axis y line=left,
               axis x line=bottom, xlabel=CLIP score $\uparrow$, 
            width=4.2in,
            height=2.2in,
            grid=major,
            legend style={at={(0.5,1.0)},anchor=north}, legend columns=2]
            \addlegendentry{SD-1.5}
             \addplot[smooth, mark=pentagon,blue!70!, mark size =2pt, very thick] plot coordinates {
             (28.7974548339843,17.1062725022054)
             (29.6813602447509,12.2676906252462)
             (30.4680709838867,9.93290931104024)
             (30.8049221038818,10.3816995867157)
             (30.9646968841552,11.2170359912854)
             (31.0656986236572,12.0411636080433)
             (31.1128349304199,12.6972582624043)
             (31.1752414703369,13.5305102478146)
             };
             \addplot[smooth, mark=*,orange, mark size =2pt, very thick] plot coordinates {
             (29.39182472229,11.7844031437169)
             (30.0867729187011,9.6722834900442)
             (30.7158946990966,9.66620283319713)
             (30.9658889770507,10.7763662676596)
             (31.138219833374,11.9259335462949)
             (31.2144985198974,12.8627415619918)
             (31.2379436492919,13.6193031540383)
             (31.3067398071289,14.2646390250415)
             };
             \addlegendentry{+ SingDiffusion}
     \end{axis}
     \end{tikzpicture}
     \hspace{7mm}
     \begin{tikzpicture}[font=\small,scale=0.8]
         \begin{axis}[
            ymin=8.5, ymax=15,
            ytick={9,10,...,14},
            xmin=29.35, xmax=31.8,
            ylabel=y, 
            ylabel=FID score $\downarrow$, y label style={at={(0.05,0.5)}},
               axis y line=left,
               axis x line=bottom, xlabel=CLIP score $\uparrow$, 
            width=4.2in,
            height=2.2in,
            grid=major,
            legend style={at={(0.5,1.0)},anchor=north}, legend columns=2]
            \addlegendentry{SD-2.0-base}
             \addplot[smooth, mark=pentagon,blue!70!, mark size =2pt, very thick] plot coordinates {
             (29.3887252807617,13.4162656986438)
             (30.1889953613281,10.0299095343251)
             (30.926902770996,9.37288781665751)
             (31.2400455474853,10.2314892721161)
             (31.4055519104003,11.378623602961)
             (31.5057258605957,12.1982183625034)
             (31.5679454803466,13.034090109423)
             (31.6097621917724,13.718830686026)
             };
             \addplot[smooth, mark=*,orange, mark size =2pt, very thick] plot coordinates {
             (29.8600921630859,10.2488082350809)
             (30.5244083404541,8.83312058323991)
             (31.1151008605957,9.57954180650335)
             (31.3914775848388,10.8908514477453)
             (31.5326499938964,12.0060631539241)
             (31.6213970184326,12.9602165936181)
             (31.6795387268066,13.6015354525134)
             (31.7311325073242,14.3445059991605)
             };
             \addlegendentry{+ SingDiffusion}
     \end{axis}
     \end{tikzpicture}
\caption{Comparison of Pareto curves between SingDiffusion, SD-1.5, and SD-2.0-base on 30k COCO images, across various guidance scales in [1.5, 2, 3, 4, 5, 6, 7, 8].}
\label{fig:curve}
\end{figure*}
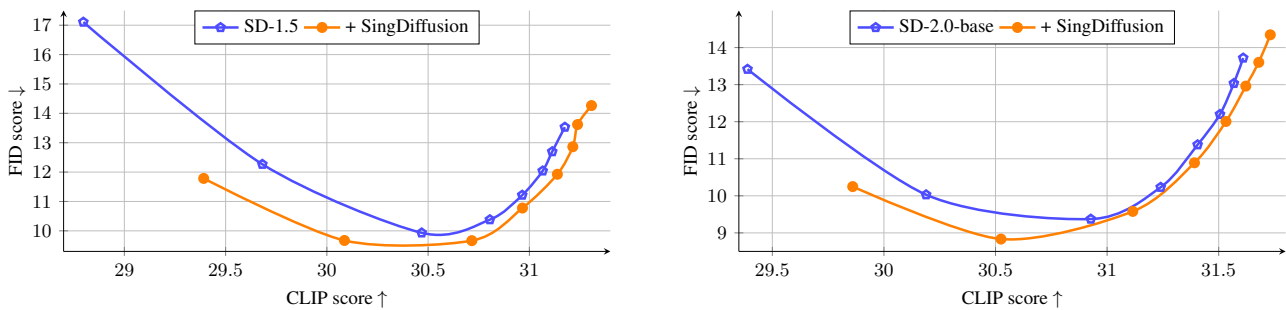

\begin{figure*}[t]
\centering
\includegraphics[width=1\textwidth]{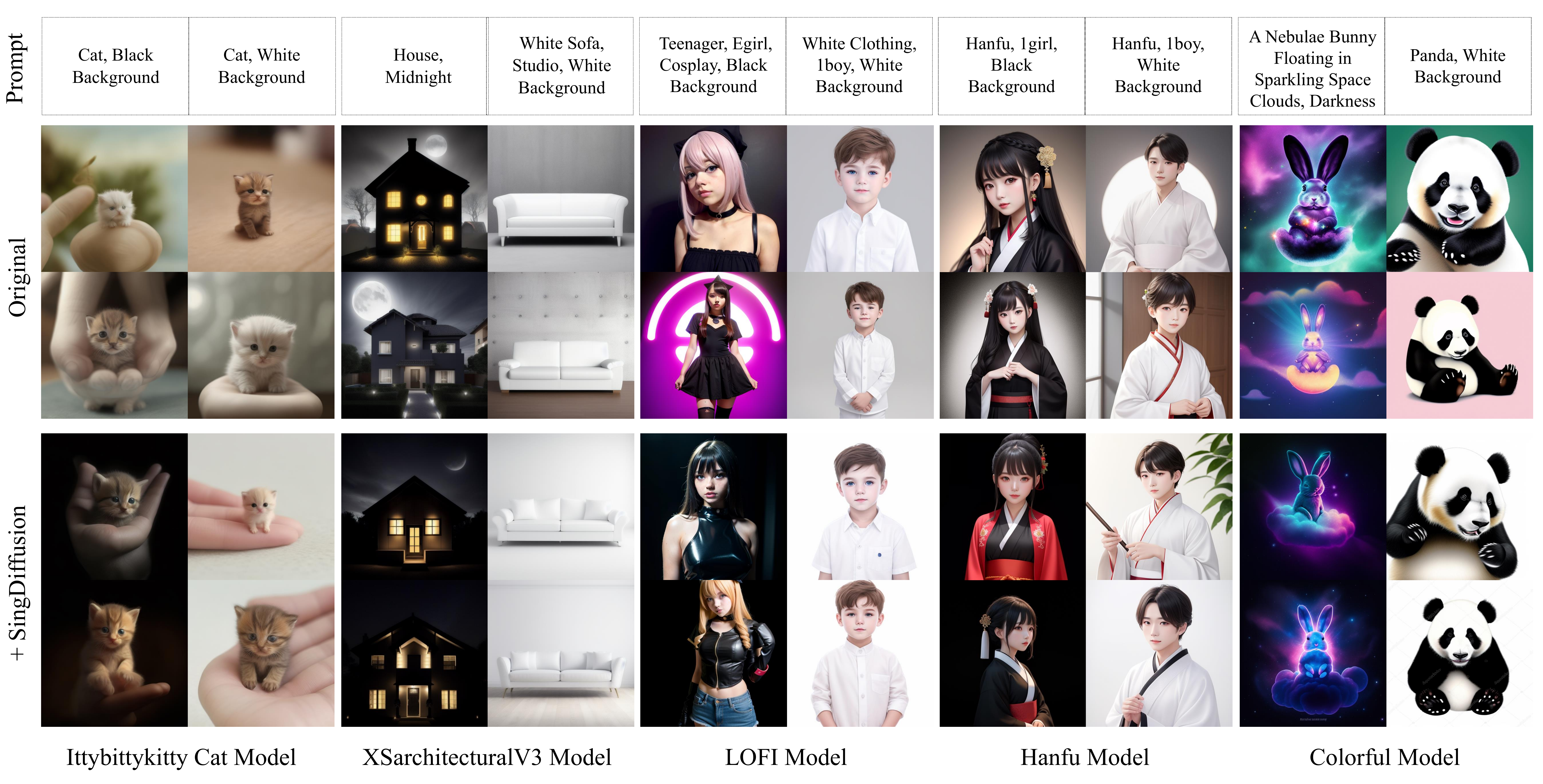} 
\caption{Our method can be trained once and seamlessly integrated into the pre-trained models on CIVITAI in a plug-and-play fashion.}
\label{fig:app}
\end{figure*}

\section{Experiment}
\subsection{Implement details}
To create a versatile plug-and-play model, our model is trained on the Laion2B-en dataset \cite{schuhmann2022laionb}, including 2.32 billion text-image pairs. 
The images are center-cropped to 512 $\times$ 512.
Our $\bar{y}_\theta$ follows the U-net structure similar to stable diffusion models but with reduced parameters, totaling 140 million. 
We utilize the AdamW optimizer \cite{loshchilov2018decoupled} with a learning rate of 1e-4 to train our $\bar{y}_\theta$. 
The training is executed across 64 Nvidia V100 chips with a batch size of 3072, and completes after 180K steps.

During the testing process, all steps, including our initial sampling, are carried out using the default guidance scale of $w=7.5$. 
After our initial sampling at the singular time step, all existing pre-trained diffusion models generate images in the DDIM pipeline, executing 50 steps to produce images following a schedule with a total time of $T = 1000$.

\setlength{\tabcolsep}{0.1mm}{
\begin{table}[t]
\centering
\renewcommand{\arraystretch}{1.1}
\caption{Comparison of average brightness of 100 generated images between stable diffusion models and our SingDiffusion under different prompt conditions. For 'white'/'black' prompts, higher/lower average brightness is better.}
\label{tab:gray}
\begin{tabular}{c|c|c|c|c}
\Xhline{1pt}
        \small Model & \begin{tabular}[c]{@{}c@{}} \scriptsize"Solid \textbf{white} \vspace{-8pt} \\ \scriptsize background"\end{tabular} & \begin{tabular}[c]{@{}c@{}}\scriptsize"Solid \textbf{black} \vspace{-8pt} \\ \scriptsize background"\end{tabular} & \begin{tabular}[c]{@{}c@{}c@{}}\scriptsize"Monochrome \scriptsize \vspace{-8pt} \\ \scriptsize line-art logo on a \vspace{-8pt} \\ \scriptsize \textbf{white} background"\end{tabular} & \begin{tabular}[c]{@{}c@{}c@{}}\scriptsize"Monochrome \scriptsize \vspace{-8pt} \\ \scriptsize line-art logo on a \vspace{-8pt} \\ \scriptsize \textbf{black} background"\end{tabular}  \\ \Xhline{1pt}
\small SD-1.5 &        141.43                  &           83.09               &                                                          137.95                                   &      113.66                                                                                       \\
\small + Ours  &         \textbf{212.59}                 &            \textbf{3.04}              &    \textbf{223.68}                                                                                         &      \textbf{11.52}                                                                                       \\ \Xhline{1pt}
\small SD-2.0-base &        150.52                  &     99.67                     &    136.13                                                                                         &     104.45                                                                                        \\ 
\small + Ours  &     \textbf{227.43}                     &         \textbf{0.29}                 &                        \textbf{228.68}                                                                     &    \textbf{10.87}                                                                                         \\ \Xhline{1pt}
\end{tabular}
\end{table}
}

\subsection{Average brightness issue}

To validate the effectiveness of SingDiffusion in addressing the average brightness issue, we select four extreme prompts, including ``Solid white/black background'', and ``Monochrome line-art logo on a white/black background''. 
For each prompt, we generate 100 images using SD-1.5, SD-2.0-base and SingDiffusion methods, and then calculate their average brightness. 
The results are shown in Table \ref{tab:gray}.
It is remarkably clear that the stable diffusion methods, whether using prompts with `black' or `white' descriptors, tend to generate images with average brightness.
However, SingDiffusion, implementing initial singularity sampling, effectively corrects the average brightness of the generated images. 
For example, under the prompt ``solid black background,'' our method notably lowers the average brightness from 99.67 to 0.29 for images generated by SD-2.0-base.

Moreover, we also provide visualization results for these prompts in Fig. \ref{fig:black}. 
It can be seen that images generated by stable diffusion methods predominantly exhibit a gray tone.
In contrast, our SingDiffusion method successfully overcomes this issue, and is capable of generating both dark and bright images.

\subsection{Improvement on image quality}

To validate our improvements in general image quality, following \cite{saharia2022photorealistic}, we randomly select 30k prompts from the COCO dataset \cite{lin2014microsoft} as the test set.
We employ two metrics for evaluation, including the FID score and CLIP score. 
Specifically, Fréchet Inception Distance (FID) \cite{heusel2017gans} calculates the Fréchet distance between the real data and the generated data.
A lower FID implies more realistic generated data.
While the Contrastive Language-Image Pre-training (CLIP) \cite{radford2021learning} score measures the similarity between the generated images and the given prompts. 
A higher CLIP score means the generated images better match the input prompts.

\setlength{\tabcolsep}{2.2mm}{
\begin{table}[!t]
\centering
\renewcommand{\arraystretch}{1.1}
\caption{Comparison of stable diffusion model and SingDiffusion on FID score and CLIP score without classifier guidance.}
\begin{tabular}{c|cc|cc}
\Xhline{1pt}
{\multirow{2}{*}{Model}} & \multicolumn{2}{c|}{SD-1.5}    & \multicolumn{2}{c}{SD-2.0-base}\\ \Xcline{2-5}{0.5pt}
                      & FID $\downarrow$  & CLIP $\uparrow$    & FID $\downarrow$  & CLIP $\uparrow$ \\ \Xhline{1pt} Original & 31.86 & 26.70 & 25.17 & 27.48  \\
+ SingDiffusion & \textbf{21.09} & \textbf{27.71}  & \textbf{18.01} & \textbf{28.23} \\ \Xhline{1pt}
\end{tabular}
\label{tab:FID}
\end{table}
}

First of all, we compare SingDiffusion with SD-1.5 and SD-2.0-base in Table \ref{tab:FID} to gauge the model's inherent fitting capability, without using guidance-free techniques.
It is evident that SingDiffusion significantly outperforms the existing stable diffusion methods in both FID and CLIP scores. 
These results highlight the yet-to-be-utilized fitting potential in current stable diffusion models, emphasizing the significance of sampling at the initial singular time step.

Furthermore, inspired by Imagen \cite{saharia2022photorealistic}, we plot CLIP v.s. FID Pareto curves by varying guidance values within the range [1.5, 2, 3, 4, 5, 6, 7, 8] in Fig. \ref{fig:curve}. 
SingDiffusion exhibits substantial improvements over stable diffusion models, especially noticeable with smaller guidance scales.
As the guidance scale increases, SingDiffusion consistently maintains a lower FID compared to stable diffusion for achieving a similar CLIP score.
This emphasizes that our approach not only enhances image realism but also ensures better adherence to the input prompts.

\subsection{Plug-and-play on other pre-trained models}

Since our method is extensively trained on the Laion dataset, it can be easily integrated into the majority of existing diffusion models. 
To validate this, we download several pre-trained models from the CIVITAI website, each specializing in different image domains like anime, animals, clothing, and so on.
To facilitate the application of SingDiffusion to these models, we integrate SingDiffusion into the popular Stable Diffusion Web UI \cite{webui} system, and sample images with the Euler ancient method after using our initial singularity sampling process.
The outcomes, displayed in Fig. \ref{fig:fig1} and Fig. \ref{fig:app}, showcase SingDiffusion effectively resolving the average brightness issue across all models while preserving their original generative capacities. 
This demonstrates the adaptability and practicality of our approach.

\begin{figure}[!t]
     \begin{tikzpicture}[font=\small,scale=0.8]
         \begin{axis}[
            ymin=9.3, ymax=16.5,
            ytick={10,...,14,15,16,17},
            xmin=29.3, xmax=31.4,
            ylabel=y, 
            ylabel=FID score $\downarrow$, y label style={at={(0.05,0.5)}},
               axis y line=left,
               axis x line=bottom, xlabel=CLIP score $\uparrow$, 
            width=4.2in,
            height=2in,
            grid=major,
            legend style={at={(0.5,1.0)},anchor=north}, legend columns=2]
            \addlegendentry{SingDiffusion w/o norm}
             \addplot[smooth, mark=pentagon,gray, mark size =2pt, very thick] plot coordinates {
             (29.4446144104003,12.118667254164)
             (30.1287441253662,10.3935204962386)
             (30.7170085906982,10.9645081319419)
             (30.9585494995117,12.1926445171212)
             (31.1085777282714,13.3839367879316)
             (31.1806755065917,14.4272003356835)
             (31.2479877471923,15.0689848783833)
             (31.2776317596435,15.8664441966888)
             };
             \addplot[smooth, mark=*,orange, mark size =2pt, very thick] plot coordinates {
             (29.39182472229,11.7844031437169)
             (30.0867729187011,9.6722834900442)
             (30.7158946990966,9.66620283319713)
             (30.9658889770507,10.7763662676596)
             (31.138219833374,11.9259335462949)
             (31.2144985198974,12.8627415619918)
             (31.2379436492919,13.6193031540383)
             (31.3067398071289,14.2646390250415)
             };
             \addlegendentry{SingDiffusion}
     \end{axis}
     \end{tikzpicture}
     \caption{Ablation study of guidance normalization on SD-1.5 across various guidance scales in [1.5, 2, 3, 4, 5, 6, 7, 8].}
     \label{fig:curve1.5}
\end{figure}
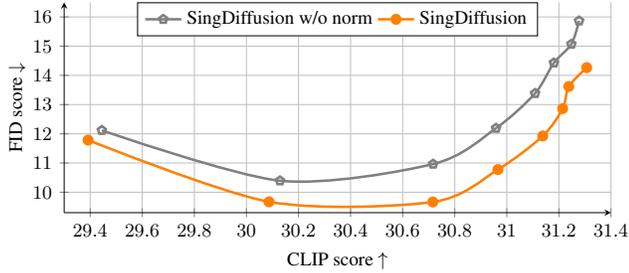

\begin{figure}[!t]
\centering
\includegraphics[width=0.48\textwidth]{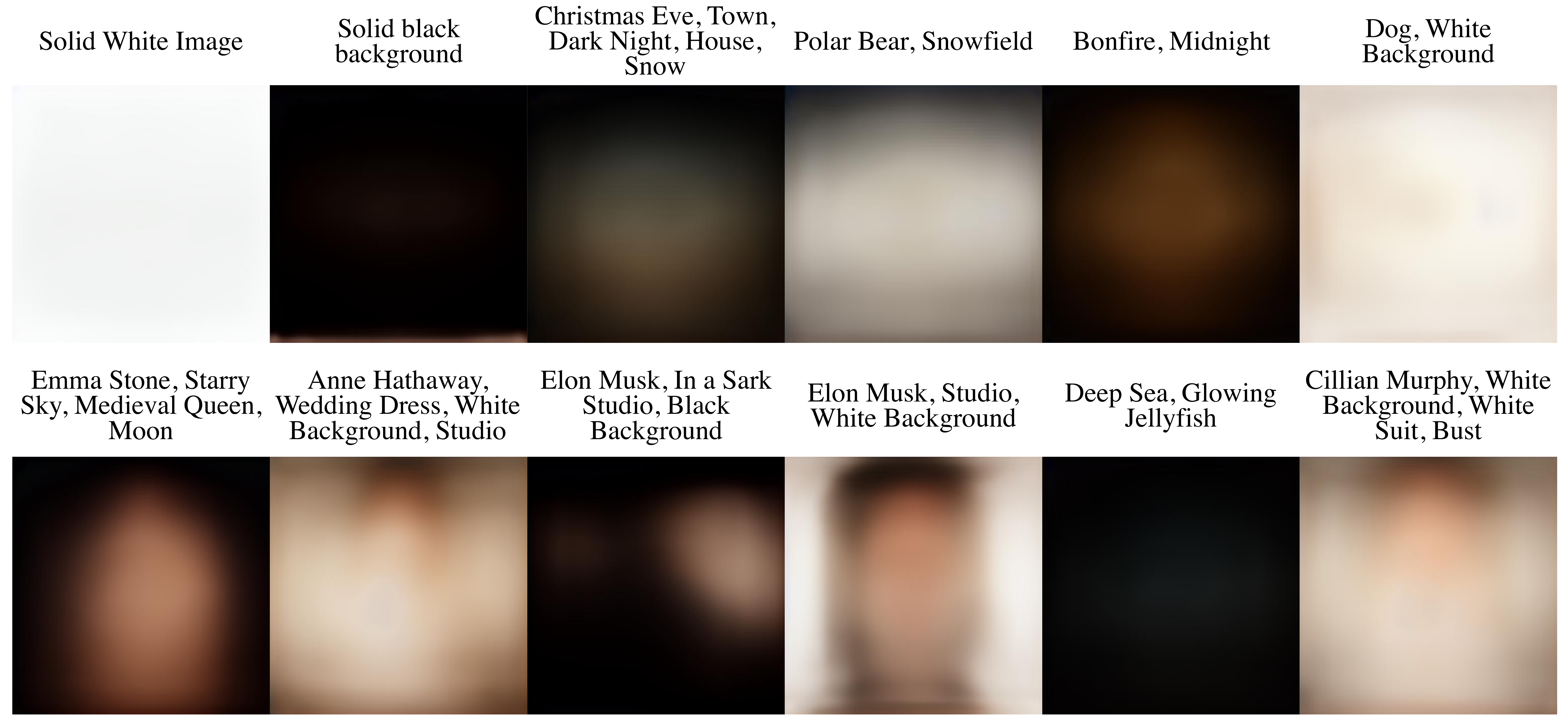} 
\caption{Visualization of ${\bar{y}_\theta}$ for various prompts.}
\label{fig:x0}
\end{figure}

\subsection{Effect on classifier guidance normalization}

We conduct an ablation study on the guidance normalization operation (Eq. \ref{eq:guidance}), and represent the CLIP v.s. FID Pareto curves in Fig. \ref{fig:curve1.5}.
It can be seen that the method without normalization exhibits inferior results compared with the full method. 
As the guidance scale increases, the gap in FID between these two methods gradually widens.
This phenomenon primarily arises from significant disparities between ${\bar{y}_{{\theta}_{pos}}}$ and ${\bar{y}_{{\theta}_{neg}}}$.
According to Eq. \ref{eq:guidance}, larger guidance scales may lead to overflow issues.
In contrast, normalizing the results with the guidance scale  helps keep the outputs within a typical range, thus restoring the FID score.

\begin{figure}[!t]
\centering
\includegraphics[width=0.48\textwidth]{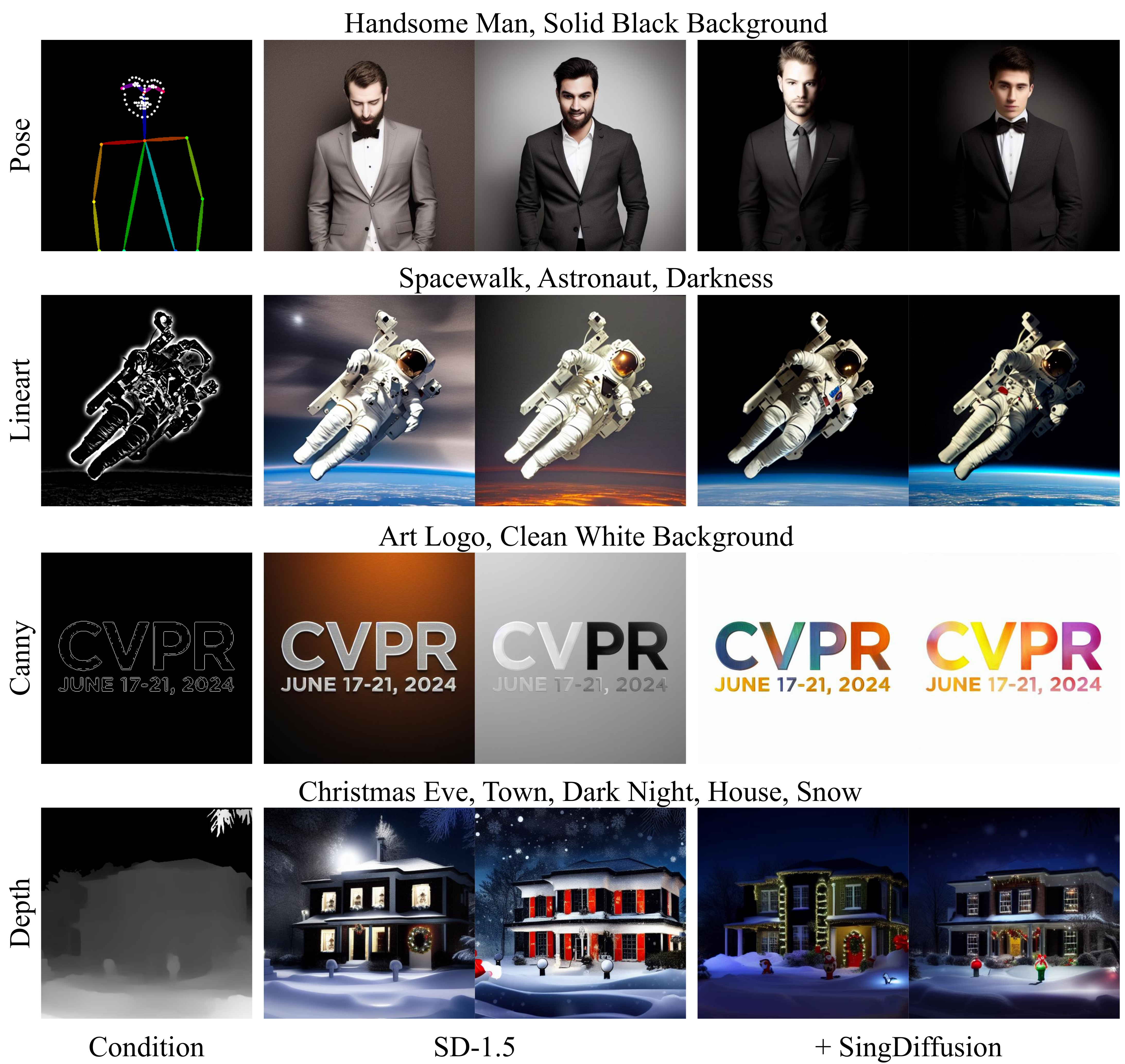} 
\caption{SingDiffusion integrates seamlessly with ControlNet.}
\label{fig:control}
\end{figure}

\subsection{Visualization of the \texorpdfstring{${\bar{y}_{\theta}}$}{\texttwoinferior}}
According to Eq. \ref{eq:loss}, our main goal with $\bar{y}_{\theta}$ is to model the average image corresponding to each prompt. 
To confirm this, we employ the prompts presented in Fig. \ref{fig:fig1} and visualize their corresponding $\bar{y}_\theta$. 
As demonstrated in Fig. \ref{fig:x0}, it is clear that $\bar{y}_\theta$ does indeed represent a smoothed average image.
For instance, the first four images of the second row resemble average faces, aligning with the input prompt for generating celebrities. 
Additionally, we notice that $\bar{y}_\theta$ adapts to the prompt's brightness suggestion, appearing brighter or darker accordingly. 
This highlights our method's ability to effectively capture pertinent lighting conditions, and underscores the significance of the initial singular time step sampling.

\subsection{Application on ControlNet}

Our model can seamlessly integrate with existing diffusion model plugins, such as ControlNet \cite{Zhang_2023_ICCV}. 
Since our $\bar{y}_{\theta}$ is structurally different from ControlNet, ControlNet is adopted after our initial sampling step.
The results in Fig. \ref{fig:control} demonstrate that our method is fully compatible with ControlNet and effectively resolves its average brightness issue.

\section{Conclusion}
In this study, we delve into the singularities of time intervals in diffusion models, exploring both theoretical and practical aspects. 
Firstly, we demonstrate that at both regular and singular time steps, the reverse process can be approximated by a Gaussian distribution. 
Leveraging these theoretical insights, we conduct an in-depth analysis of singular time step sampling and propose a theoretical solution. 
Finally, we introduce a novel plug-and-play method SingDiffusion, addressing the initial time-step sampling challenge. 
Remarkably, this module substantially mitigates the average brightness issue prevalent in most current diffusion models, and also enhances their generative capabilities.

\paragraph{Acknowledgments.} This project is supported by the National Natural Science Foundation of China (62072482, U22A2095)

{
    \small  \bibliographystyle{ieeenat_fullname}
    \bibliography{main}
}

\clearpage
\setcounter{page}{1}
\maketitlesupplementary
\appendix

\section{Singularities for SDE}
\label{sec:singu_sde}

In this section, we'll address the singularities within the continuous diffusion process of the SDE.
The SDE is represented by the Ornstein-Uhlenbeck equation:
\begin{equation} \label{eq:ou}
    d x_t = f(t) x_t dt + g(t) d B_t,
\end{equation}
where 
$f(t)$ is the drift coefficient, $g(t)$ is the diffusion coefficient, and $B_t$ is the Brownian Motion. The solution to Eq. \ref{eq:ou} is:
\begin{equation}\label{eq:ou_sol}
\begin{aligned}
    x_t =&\exp (\int_s^t f(\tau) d\tau) x_s \!+ \! \int_s^t g(\nu) \exp (-\!\int_\nu^t f(\tau) d\tau) dB_\nu \\
    =& \frac{\alpha_t}{\alpha_s} x_s + (\int_s^t g^2(\nu) \frac{\alpha_t^2}{\alpha_\nu^2} d \nu)^\frac{1}{2} B_1,
\end{aligned}
\end{equation}
where $\alpha_t = \exp (\int_0^t f(\tau) d\tau)$. In order to maintain a constant variance and achieve convergence to a standard Gaussian distribution, it is necessary to satisfy the conditions of $f(t) = -\frac{1}{2}g^2(t)$ and $\lim_{t \to 1} f(t) = -\infty$. By fulfilling these requirements, Eq. \ref{eq:ou_sol} can be rewritten as follows:
\begin{equation}\label{eq:ou_sol_simp}
    x_t = \alpha_{t|s} x_s + \sigma_{t|s} B_1,
\end{equation}
where $\alpha_{t|s} = \alpha_{t}/\alpha_{s}$, $\sigma_{t|s} = \sqrt{1-\alpha_{t|s}^{2}}$. The single-time marginal distribution $p(x_t, t)$ of this process remains consistent with Eq. \ref{stm}, while the forward conditional distribution $p(x_t | x_s)$ is given by:
\begin{equation} \label{eq:forward_cond}
\begin{aligned}
    p(x_t | x_s) = (2 \pi \sigma_{t|s}^2)^{-\frac{1}{2}} \exp(-\frac{(x_t - \alpha_{t|s} x_s) ^ 2}{2 \sigma_{t|s}^2}).
\end{aligned}
\end{equation}
According to \cite{song2020score,anderson1982reverse}, the reverse-time SDE is:
\begin{equation} \label{eq:rev_sde}
    d x_t = (f(t)x_t - g^2(t) \nabla_{x_t} p(x_t, t)) dt + g(t) d \tilde{B}_t,
\end{equation}
where $\tilde{B}_t$ is a reverse time Brownian motion. The term $\nabla_{x_t} p(x_t, t)$, is metiond as the score function:
\begin{equation}
\begin{aligned}
 s(x_t, t):=\nabla_{x_t} p(x_t, t) = \sum_i w_i(x_t) \frac{\alpha_t y_i - x_t}{\sigma_t^2},
\end{aligned}
\end{equation}
where $w_i(x_t, t) = \frac{\exp(-\frac{(x_t - \alpha_t y_i)^2}{2 \sigma_t^2})}{\sum_j \exp(-\frac{(x_t - \alpha_t y_j)^2}{2 \sigma_t^2})}$. This score function can be fitted by a neural network $s_\theta(x_t, t)$ using score matching.

\subsection{The singularity at $t=1$}

\subsubsection{Singularity in SDE Solution}
To solve the SDE, the initial step involves rewriting the Eq. \ref{eq:ou} as follows:
\begin{equation}
\begin{aligned}
    d \frac{x_t}{\alpha_t} &= -f(t) \frac{x_t}{\alpha_t} dt + \frac{1}{\alpha_t} (f(t)x_t dt + g(t) d B_t) \\
    &=\frac{g(t)}{\alpha_t} d B_t.
\end{aligned}
\end{equation}

However, it's crucial to note that a singularity occurs when $\alpha_1 = 0$, leading to a division-by-zero issue. Consequently, Eq. \ref{eq:ou_sol_simp} and Eq. \ref{eq:forward_cond} may not hold at $t=1$. Nonetheless, these equations remain valid for all $s \in [0, 1)$, allowing us to deduce the following:
\begin{Proposition}
\label{mesure_converge_cond}
    $\forall 0 \leq s < t \leq 1$, the probability density $p(x_t | x_s)$ converges  to $\mathcal{N}(\bm{0},\bm{I})$ point-wisely, as $t \to 1$.
\end{Proposition}

\begin{Corollary}
\label{mesure_converge}
    The probability density $p(x_t, t)$ converges to $\mathcal{N}(\bm{0},\bm{I})$ point-wisely, as $t \to 1$.
\end{Corollary}

These claims demonstrate that the Eq. \ref{eq:ou_sol_simp} and Eq. \ref{eq:forward_cond} can be extended to the interval $[0, 1]$.

\subsubsection{Singularity in Kolmogorov Equations}

According to \cite{anderson1982reverse}, the reverse process in Eq. \ref{eq:rev_sde} relies on the Kolmogorov forward equation and backward equation:
\begin{equation} \label{eq:kolmogorov}
\begin{aligned}
    &\frac{\partial{p(x_t | x_s)}}{\partial{t}} = - f(t) \nabla_{x_t} (x_t p(x_t | x_s)) + \frac{1}{2} g^2(t) \Delta p(x_t | x_s), \\
    &\frac{\partial{p(x_t | x_s)}}{\partial{s}} = - f(t) x_t \nabla_{x_t} p(x_t | x_s) - \frac{1}{2} g^2(t) \Delta p(x_t | x_s).
\end{aligned}
\end{equation}
However, as $-f(1) = \frac{1}{2} g^2(1) = \infty$, these equations may not hold at $t=1$. As a result, the reverse process (\ref{eq:rev_sde}) may not establish at $t=1$.

\subsubsection{Singularity in Discretization}

With the trained network $s_\theta(x_t)$, \cite{song2020score} introduce a sampling technique referred to as \emph{reverse diffusion sampling}. It discretizes the Eq. \ref{eq:rev_sde} using the Euler–Maruyama method:
\begin{equation} \label{eq:rev_sde_dis}
\begin{aligned}
    x_s = x_t - (f(t) x_t - g^2(t) s_\theta(x_t)) (t - s) + g(t) \sqrt{t - s} z_t.
\end{aligned}
\end{equation}
As $-f(1) = \frac{1}{2} g^2(1) = \infty$ and $s_\theta(x_1) \approx x_1$, the right-hand side of this equation becomes $\infty - \infty$ at $t = 1$. This observation highlights the limitation of naively discretizing Eq. \ref{eq:rev_sde} near $t=1$.

\subsubsection{Tackling the Singularity}

To mitigate the singularity challenge in the Kolmogorov equations and the reverse SDE, we scrutinize the reverse conditional distribution at $t = 1$, denoted as $p(x_s | x_1)$. 
Given that Proposition \ref{mesure_converge_cond} and Corollary \ref{mesure_converge} affirm the equivalence of both $p(x_1, 1)$ and $p(x_1 | x_s)$, following a standard Gaussian distribution, we deduce:
\begin{equation} \label{eq:sde_bayes}
    p(x_s | x_1) = p(x_1 | x_s) \frac{p(x_s, s)}{p(x_1, 1)} = p(x_s, s),
\end{equation}
which holds for all $s \in [0, 1]$. Therefore, we successfully solved the reverse equation at $t=1$, circumventing the Kolmogorov equations.

We now possess the necessary tools to overcome the singularity present in Eq. \ref{eq:rev_sde_dis}. By setting $t=1$ in proposition 2, we obtain the following corollary:

\begin{Corollary}
    $\exists \tau \in [0, 1)$ and $C > 0$, such that $\forall t \in [\tau ,1]$, $\int_{\mathbb{R}^d} |p(x_t, t) - \tilde{p}(x_t, t)| d x_t < C \sqrt{\alpha_t}$, where
    \begin{equation}\label{eq:prop5}
    \begin{aligned}
        \tilde{p}(x_t, t) = (2 \pi \sigma_t^2)^{-\frac{1}{2}} \exp (- \frac{(x_t - \alpha_t \bar{y}(x_t, t))^2}{2 \sigma_t^2})
    \end{aligned}
    \end{equation}
\end{Corollary}

As a result, at $t=1$,  we're able to generate a step by employing Eq. \ref{eq:sde_bayes} and Eq. \ref{eq:prop5},  followed by sampling the subsequent steps utilizing the reverse diffusion sampling method outlined in Eq. \ref{eq:rev_sde_dis}. Remarkably, this process aligns with the approach we employ to tackle the singularity in DDPM.

It is worth to note that Eq. \ref{eq:sde_bayes} also offers additional information as follows:
\begin{Proposition}
    The generated sample $x_0$ is independent with the initial noise point $x_1$
\end{Proposition}
Consider the \emph{sample space} as the set formed by the diffusion model's outputs. Within this framework, we can assert the following corollary:

\begin{Corollary}
    Given any fixed initial noise point ${x_1}_{fix}$, the entire sample space can be generated with the reverse diffusion sampling method, or equivalently, the DDPM sampling method.
\end{Corollary}

\subsection{The singularity at $t=0$}
\label{sec:sde_t0}

Similar to DDPM, the SDE also encounters singularities at $t=0$. These singularities stem from the score function:
\begin{equation}
\begin{aligned}
    s(x_t, t) &= \nabla_{x_t} \log p(x_t, t) = - \sum_i w_i(x_t) \frac{x_t - \alpha_t y_i}{\sigma_t^2} \\
    &= - \frac{x_t - \alpha_t \sum_i w_i(x_t) y_i}{\sigma_t^2} = - \frac{x_t - \alpha_t \bar{y}(x_t, t)}{\sigma_t^2}.
\end{aligned}
\end{equation}
Let $j_0 = \mathop{\arg\min}\limits_{j} |x_t - \alpha_t y_j|$, we have
\begin{equation}\label{score_singu}
\begin{aligned}
    &\lim_{t \to 0^+} -\sum_i w_i(x_t) \frac{(x_t - \alpha_t y_i)}{\sigma_t^2} \\
    & = \lim_{t \to 0^+} \sum_i \frac{-\frac{(x_t - \alpha_t y_i)}{\sigma_t^2}}{\sum_j \exp \frac{1}{2\sigma_t^2}((x_t - \alpha_t y_i)^2 - (x_t - \alpha_t y_j)^2)} \\
    & = \begin{cases}
        0, & \text{if}~~x_{t=0}=y_{j_0},\\
        (x_{j_0} - x_{t=0}) * \infty, & \text{if}~~x_{t=0}\neq y_{j_0},
    \end{cases}
\end{aligned}
\end{equation}
where $(x_{j_0} - x_{t=0}) * \infty$ signifies that the vector is oriented in the direction of $x_{j_0} - x_{t=0}$ and possesses an infinite norm.
This Equation demonstrates that as $t$ approaches 0, $x_t$ is driven towards the nearest training sample. 
This aligns with the discrete scenario depicted in Eq. \ref{dis_t1}. 
Despite the presence of singularities in the score function, it guides the sampling process toward convergence at a specific point. 
Consequently, the behavior near $t=0$ becomes predictable, indicating that the singularities do not need avoidance if we employ suitable sampling techniques.
Similar to DDPM, the final sampling step is $x_0 = \Bar{y}(x_t, t)$.

\section{Singularities for Probability Flow ODE}
\label{sec:singu_ode}

\subsection{Singularites for the Reverse Process}\label{sec:singODE}

The probability flow is defined by \cite{song2020score}:
\begin{equation} \label{eq:prob_ode}
    d x_t = (f(t) x_t - \frac{1}{2} g^2(t) s(x_t, t)) dt.
\end{equation}
Expanding the expression for $s(x_t, t)$, we obtain:
\begin{equation} \label{eq:prob_ode1}
    \frac{d x_t}{dt} = -\frac{\alpha_t \alpha_t'}{1 - \alpha_t^2} x_t + \frac{\alpha_t'}{1 - \alpha_t^2} \bar{y}(x_t, t),
\end{equation}
where $\alpha_t'$ represents the time derivative of $\alpha_t$. In practical implementations, the design of $\alpha_t$, such as the cosine function, linear function, etc., does not result in an infinite derivative at $t=1$.
Consequently, there are no singularities present at $t=1$ by using $x$-prediction for the ODE.

At $t=0$, the score function will drive $x_t$ towards the nearest training sample, which aligns with the behavior described in section \ref{sec:sde_t0}.

\subsection{The score function can not be Lipschitz continuous at $t=0$}

Several previous papers have mentioned the Lipschitz problem of the ODE near $t=0$. However, it is important to note that this behavior is not a ``problem" but rather a characteristic of the ODE.

We can disprove the assumption of the ODE's Lipschitz continuity at $t=0$ through a counter-proof. 
As $s(x_t, t)$ is Lipschitz continuous in $[\epsilon,1], \forall \epsilon>0$, if we assume the ODE to be Lipschitz continuous near $t=0$, then it will be continuous throughout the interval $[0,1]$. 
According to the Picard-Lindel\"{o}f theorem, the solution $x_t$ remains unique on $[0,1]$ for any initial condition. 
Consequently, the single-time marginal distribution $p(x_t, t)$ consistently comprises a sum of Dirac deltas, given the initial distribution (Eq. \ref{init_dist}) as a sum of Dirac deltas.  
However, this leads to a discrepancy with the expected form in Eq. \ref{stm}, resulting in a contradiction.
Consequently, the right-hand side of Eq. \ref{eq:prob_ode} cannot exhibit Lipschitz continuity near $t=0$. This leads us to the following proposition:

\begin{Proposition}
    The right-hand side of Eq. \ref{eq:prob_ode} and the score function $s(x_t, t)$ cannot be Lipschitz continuous across the space $\mathbb{R}^d \times [0,1]$.
\end{Proposition}

The absence of Lipschitz continuity implies the potential existence of multiple solutions for a given initial condition. 
These solutions must adhere to specific distributions to ensure the single-time marginal distribution in Eq. \ref{stm}.
Consequently, in the reverse process, different $x_t$ values will converge to the points within the training samples  $\{ y_i \}_{i=1}^N$.

\section{Singularities for DDIM}
\label{sec:singu_ddim}

Since DDIM is a discretization of the ODE \cite{song2020denoising}, according to Section \ref{sec:singODE}, there are no singularities by using $x$-prediction at $t=1$. We can obtain DDIM sampling by discretizing Eq. \ref{eq:prob_ode1} as follows:
\begin{equation}
\begin{aligned}
    x_t - x_s = &-\frac{\alpha_t}{\sigma_t^2} (\alpha_t - \alpha_s) x_t + \frac{\alpha_t - \alpha_s}{\sigma_t^2} \bar{y}(x_t, t) \\
    &+ o(|\alpha_t - \alpha_s|),
\end{aligned}
\end{equation}
Rearranging the formula, we have:
\begin{equation}
\begin{aligned}
    x_s &= \frac{\alpha_s - \alpha_t}{\sigma_t^2} \bar{y}(x_t, t) + \frac{1 - \alpha_s \alpha_t}{\sigma_t^2} x_t + o(|\alpha_t - \alpha_s|) \\
    &= (\alpha_s - \alpha_t \frac{\sqrt{\sigma_s^2\sigma_t^2 + (\alpha_t - \alpha_s)^2}}{\sigma_t^2}) \bar{y}(x_t, t)\\
    &~~~~+ \frac{\sqrt{\sigma_s^2\sigma_t^2 + (\alpha_t - \alpha_s)^2}}{\sigma_t^2} x_t + o(|\alpha_t - \alpha_s|) \\
    & = (\alpha_s - \frac{\sigma_s}{\sigma_t} \alpha_t) \bar{y}(x_t, t) + \frac{\sigma_s}{\sigma_t} x_t + o(|\alpha_t - \alpha_s|).
\end{aligned}
\end{equation}
In Algorithm \ref{alg:sampling}, we utilize DDIM to sample the initial step of the reverse process.

\section{Proofs}
\label{sec:proof}

\begin{Lemma}
    $\forall ~ 1 \leq i \leq N$ and $t \in [0, 1]$, there is a constant $M$ such that
    \begin{equation}
        |y_i - \bar{y}(x_t, t)| \leq M
    \end{equation}
    and
    \begin{equation}
        |\bar{y}(x_t, t)| \leq M.
    \end{equation}
\end{Lemma}

\paragraph{Proof:}
Let $M_1 = \max \{|y_i - y_j|: 1 \leq i,j \leq N \}$ and $M_2 = \max \{ |y_i| \}$. 
\begin{equation}
\begin{aligned}
    |y_i - \bar{y}(x_t, t)| \leq \sum_j w_j(x_t, t) |y_i - y_j| \leq M_1,
\end{aligned}
\end{equation} 
since $\sum_i w_i(x_t, t) = 1$. And then
\begin{equation}
    |\bar{y}(x_t, t)| \leq M_1 + \max \{ |y_i| \} = M_1 + M_2.
\end{equation}
$M = M_1 + M_2$ is the constant.  \hfill $\square$

\begin{Lemma}
    $\forall 0 \leq s < t \leq 1 $ and $s$ fixed, $\sigma_{s|t}$ is non-decreasing with respect to $t$.
\end{Lemma}

\paragraph{Proof:}
\begin{equation}
\begin{aligned}
    \sigma_{s | t}^2 =& (1 - \frac{\alpha_t^2}{\alpha_s^2}) \frac{1 - \alpha_s^2}{1 - \alpha_t^2} \\
    =& \frac{1 - \alpha_s^2}{\alpha_s^2} (1 + \frac{1 - \alpha_s^2}{\alpha_t^2 - 1}).
\end{aligned}
\end{equation}
Since $\alpha_t \in [0, 1)$, $\sigma_{s|t}$ is non-increasing with respect to $\alpha_t^2$.
As $\alpha_t^2$ is non-increasing with respect to $t$, the claim is established.
In addition, the range of $\sigma_{s|t}$ is $[0,\sigma_s]$  \hfill $\square$

\begin{Lemma}
    Let $M$ be the constant in Lemma 1, $\forall s \in (0, 1)$

    (1). there is a $\tau_1$ such that $\forall t \in (s, \tau_1]$ and $1 \leq i \leq N$
    \begin{equation}
        |x_s - \frac{\alpha_{t|s} \sigma_s^2 x_t}{\sigma_t^2} - \frac{\alpha_s \sigma_{s|t}^2 \bar{y}(x_t, t)}{\sigma_s^2}| \leq \sqrt{\sigma_{s|t}}
    \end{equation}
    indicates
    \begin{equation}
        |x_s - \frac{\alpha_{t|s} \sigma_s^2 x_t}{\sigma_t^2}| \leq 2\sqrt{\sigma_{s|t}}
    \end{equation}

    (2). there is a $\tau_2$ such that $\forall t \in (s, \tau_2]$ and $1 \leq i \leq N$
    \begin{equation}
        |x_s - \frac{\alpha_{t|s} \sigma_s^2 x_t}{\sigma_t^2} - \frac{\alpha_s \sigma_{s|t}^2 \bar{y}(x_t, t)}{\sigma_s^2}| > \sqrt{\sigma_{s|t}}
    \end{equation}
    indicates
    \begin{equation}
        |x_s - \frac{\alpha_{t|s} \sigma_s^2 x_t}{\sigma_t^2} - \frac{\alpha_s \sigma_{s|t}^2 y_i}{\sigma_s^2}| > \frac{\sqrt{\sigma_{s|t}}}{2}
    \end{equation}
\end{Lemma}

\paragraph{Proof:} (1)
Let $\sigma_{s|\tau_1} = min((\frac{\sigma_s^2}{\alpha_s M})^\frac{2}{3},0.9\sigma_s)$,
then $s < t \leq \tau_1$ indicates $\frac{\alpha_s \sigma_{s|t}^2}{\sigma_s^2} M \leq \sqrt{\sigma_{s|t}}$.
Thus
\begin{equation}
\begin{aligned}
    &|x_s - \frac{\alpha_{t|s} \sigma_s^2 x_t}{\sigma_t^2}| \\
    &\leq |x_s - \frac{\alpha_{t|s} \sigma_s^2 x_t}{\sigma_t^2} - \frac{\alpha_s \sigma_{s|t}^2 \bar{y}(x_t, t)}{\sigma_s^2}| + \frac{\alpha_s \sigma_{s|t}^2}{\sigma_s^2} |\bar{y}(x_t,t )| \\
    &\leq \sqrt{\sigma_{s|t}} + \frac{\alpha_s \sigma_{s|t}^2}{\sigma_s^2} M \\
    &\leq 2\sqrt{\sigma_{s|t}}.
\end{aligned}
\end{equation}

(2) Let $\sigma_{s|\tau_2} = min((\frac{\sigma_s^2}{2 \alpha_s M})^\frac{2}{3},0.9\sigma_s)$,
then $s \leq t \leq \tau_2$ indicates $\frac{\alpha_s \sigma_{s|t}^2}{\sigma_s^2} M \leq \frac{\sqrt{\sigma_{s|t}}}{2}$.
Thus
\begin{equation}
\begin{aligned}
    &|x_s - \frac{\alpha_{t|s} \sigma_s^2 x_t}{\sigma_t^2} - \frac{\alpha_s \sigma_{s|t}^2 y_i}{\sigma_s^2}| \\
    &\geq |x_s - \frac{\alpha_{t|s} \sigma_s^2 x_t}{\sigma_t^2} - \frac{\alpha_s \sigma_{s|t}^2 \bar{y}(x_t, t)}{\sigma_s^2}| - \frac{\alpha_s \sigma_{s|t}^2}{\sigma_s^2} |y_i - \bar{y}(x_t,t )| \\
    & > \sqrt{\sigma_{s|t}} - \frac{\alpha_s \sigma_{s|t}^2}{\sigma_s^2} M \\
    & > \frac{\sqrt{\sigma_{s|t}}}{2}.
\end{aligned}
\end{equation}

\begin{Lemma}
    Let $M$ be the constant in Lemma 1

    (1). there is a $\nu_1$ such that $ \nu_1 \leq s < t \leq 1 $ and $1 \leq i \leq N$
    \begin{equation}
        |x_s - \frac{\alpha_{t|s} \sigma_s^2 x_t}{\sigma_t^2} - \frac{\alpha_s \sigma_{s|t}^2 \bar{y}(x_t, t)}{\sigma_s^2}| \leq \frac{1}{\sqrt{\alpha_{s}}}
    \end{equation}
    indicates
    \begin{equation}
        |x_s - \frac{\alpha_{t|s} \sigma_s^2 x_t}{\sigma_t^2}| \leq \frac{2}{\sqrt{\alpha_{s}}}
    \end{equation}

    (2). there is a $\nu_2$ such that $ \nu_2 \leq s < t \leq 1 $ and $1 \leq i \leq N$
    \begin{equation}
        |x_s - \frac{\alpha_{t|s} \sigma_s^2 x_t}{\sigma_t^2} - \frac{\alpha_s \sigma_{s|t}^2 \bar{y}(x_t, t)}{\sigma_s^2}| > \frac{1}{\sqrt{\alpha_{s}}}
    \end{equation}
    indicates
    \begin{equation}
        |x_s - \frac{\alpha_{t|s} \sigma_s^2 x_t}{\sigma_t^2} - \frac{\alpha_s \sigma_{s|t}^2 y_i}{\sigma_s^2}| > \frac{1}{2\sqrt{\alpha_{s}}}
    \end{equation}
\end{Lemma}

\paragraph{Proof:} (1) Let $\alpha_{\nu_1} = min((\frac{1}{M})^{\frac{2}{3}}, 0.9)$,
then $ \nu_1 \leq s < t \leq 1 $ indicates $\frac{\alpha_s \sigma_{s|t}^2 M}{\sigma_s^2} \leq \alpha_s M \leq \frac{1}{\sqrt{\alpha_s}}$.
Thus
\begin{equation}
\begin{aligned}
    &|x_s - \frac{\alpha_{t|s} \sigma_s^2 x_t}{\sigma_t^2}| \\
    & \leq |x_s - \frac{\alpha_{t|s} \sigma_s^2 x_t}{\sigma_t^2} - \frac{\alpha_s \sigma_{s|t}^2 \bar{y}(x_t, t)}{\sigma_s^2}| + |\frac{\alpha_s \sigma_{s|t}^2 \bar{y}(x_t, t)}{\sigma_s^2}| \\
    & \leq \frac{1}{\sqrt{\alpha_{s}}} + \frac{1}{\sqrt{\alpha_{s}}} = \frac{2}{\sqrt{\alpha_{s}}}.
\end{aligned}
\end{equation}

(2) Let $\alpha_{\nu_2} = min((\frac{1}{2M})^\frac{2}{3}, 0.9)$,
then $ \nu_2 \leq s < t \leq 1 $ indicates $\frac{\alpha_s \sigma_{s|t}^2 }{\sigma_s^2} M \leq \alpha_s M \leq \frac{1}{2 \sqrt{\alpha_s}}$.
Thus
\begin{equation}
    \begin{aligned}
        &|x_s - \frac{\alpha_{t|s} \sigma_s^2 x_t}{\sigma_t^2} - \frac{\alpha_s \sigma_{s|t}^2 y_i}{\sigma_s^2}| \\
        & \geq |x_s - \frac{\alpha_{t|s} \sigma_s^2 x_t}{\sigma_t^2} - \frac{\alpha_s \sigma_{s|t}^2 \bar{y}(x_t, t)}{\sigma_s^2}| \\
        & ~~~~- |\frac{\alpha_s \sigma_{s|t}^2 }{\sigma_s^2} (\bar{y}(x_t, t) - y_i)| \\
        & > \frac{1}{\sqrt{\alpha_{s}}} - \frac{1}{2\sqrt{\alpha_{s}}} = \frac{1}{2\sqrt{\alpha_{s}}}.
    \end{aligned}
    \end{equation}

\begin{Proposition_proof}
    $\forall s \in (0, 1)$, $\exists \tau \in (s, 1)$ and $C > 0$, such that $\forall t \in (s, \tau]$,
 $\int_{\mathbb{R}^{d}} |p(x_s| x_t) - \tilde{p}(x_s| x_t)| d x_s < C \sqrt{\sigma_{s|t}}$.
\end{Proposition_proof}

\paragraph{Proof:}
Let $A(x_t, y, c)$ denote the set $\{x_s : |x_s - \frac{\alpha_{t|s} \sigma_s^2 x_t}{\sigma_t^2} - \frac{\alpha_s \sigma_{s|t}^2 y}{\sigma_s^2}| > c \sqrt{\sigma_{s|t}}\}$.
\begin{equation}
\begin{aligned}
    &\tilde{P}(A(x_t, \bar{y}(x_t, t), 1) | x_t) \\
    &= \tilde{P}(|x_s - \frac{\alpha_{t|s} \sigma_s^2 x_t}{\sigma_t^2} - \frac{\alpha_s \sigma_{s|t}^2 \bar{y}(x_t, t)}{\sigma_s^2}| > \sqrt{\sigma_{s|t}} | x_t) \\
    &\leq \frac{\sigma_{s|t}^2}{\sigma_{s|t}} = \sigma_{s|t}.
\end{aligned}
\end{equation}
The last inequality is from the Chebyshev inequality.
Let $\tau_2$ be the value in Lemma 3(2), then $\forall t \in (s, \tau_2]$ $x_s \in A(x_t, \bar{y}(x_t, t), 1)$ indicates $x_s \in A(x_t, y_i, \frac{1}{2}), 1 \leq i \leq N$,
then
\begin{equation}
\begin{aligned}
    &P(A(x_t, \bar{y}(x_t, t), 1) | x_t, y_i) \\
    &\leq P(A(x_t, y_i, \frac{1}{2}) | x_t, y_i) \\
    &= P(|x_s - \frac{\alpha_{t|s} \sigma_s^2 x_t}{\sigma_t^2} - \frac{\alpha_s \sigma_{s|t}^2 y_i}{\sigma_s^2}| > \frac{1}{2} \sqrt{\sigma_{s|t}} | x_t, y_i) \\
    &\leq \frac{\sigma_{s|t}^2}{\frac{1}{4} \sigma_{s|t}} = 4\sigma_{s|t}.
\end{aligned}
\end{equation}
As a result
\begin{equation}
\begin{aligned}
    &\int_{A(x_t, \bar{y}(x_t, t), 1)} |p(x_s| x_t) - \tilde{p}(x_s| x_t)| d x_s \\
    &\leq \sum_i \int_{A(x_t, \bar{y}(x_t, t), 1)} w_i(x_t, t)|p(x_s| x_t, y_i) - \tilde{p}(x_s| x_t)| d x_s \\
    &\leq \sum_i \int_{A(x_t, \bar{y}(x_t, t), 1)} w_i(x_t, t)(p(x_s| x_t, y_i) + \tilde{p}(x_s| x_t)) d x_s \\
    &\leq \sum_i  w_i(x_t, t) 5 \sigma_{s|t} = 5\sigma_{s|t}.
\end{aligned}
\end{equation}
On the other hand, if $x_s \in A^c (x_t, \bar{y}(x_t, t), 1)$, where $A^c (x_t, y, c)$ is the complement of $A(x_t, y, c)$ denoting the set $\{x_s : |x_s - \frac{\alpha_{t|s} \sigma_s^2 x_t}{\sigma_t^2} - \frac{\alpha_s \sigma_{s|t}^2 y}{\sigma_s^2}| \leq c \sqrt{\sigma_{s|t}}\}$.
Let $\tau_1$ be the value in Lemma 3(1), $\forall t \in (s, \tau_1]$,
then $x_s \in A^c (x_t, \bar{y}(x_t, t), 1)$ indicates $x_s \in A^c (x_t, 0, 2)$.
Denoting
\begin{equation}
\begin{aligned}
    C_i &= |x_s - \frac{\alpha_{t|s} \sigma_s^2 x_t}{\sigma_t^2} - \frac{\alpha_s \sigma_{s|t}^2 \bar{y}(x_t, t)}{\sigma_s^2}| ^ 2 \\
    & ~~~~- |x_s - \frac{\alpha_{t|s} \sigma_s^2 x_t}{\sigma_t^2} - \frac{\alpha_s \sigma_{s|t}^2 y_i}{\sigma_s^2}|^2,
\end{aligned}
\end{equation}
we have
\begin{equation}
\begin{aligned}
    |\frac{C_i}{2 \sigma_{s|t}^2}| &= |\frac{1}{2 \sigma_{s|t}^2} (|x_s - \frac{\alpha_{t|s} \sigma_s^2 x_t}{\sigma_t^2} - \frac{\alpha_s \sigma_{s|t}^2 \bar{y}(x_t, t)}{\sigma_s^2}| ^ 2 \\
    & ~~~~- |x_s - \frac{\alpha_{t|s} \sigma_s^2 x_t}{\sigma_t^2} - \frac{\alpha_s \sigma_{s|t}^2 y_i}{\sigma_s^2}|^2)| \\
    & = |\frac{\alpha_s }{\sigma_s^2} (x_s - \frac{\alpha_{t|s} \sigma_s^2 x_t}{\sigma_t^2})^T (y_i - \bar{y}(x_t, t)) \\
    & ~~~~+ \frac{\alpha_s^2 \sigma_{s|t}^2}{2\sigma_s^4} (\bar{y}^2(x_t,t) - y_i^2)| \\
    & \leq \frac{\alpha_s}{\sigma_s^2} 2 \sqrt{\sigma_{s|t}} M +  \frac{\alpha_s^2 \sigma_{s|t}^2}{2\sigma_s^4} 2M^2
\end{aligned}
\end{equation}
then
\begin{equation}
\begin{aligned}
    \exp&(\frac{C_i}{2 \sigma_{s|t}^2}) = \exp( \frac{\alpha_s }{\sigma_s^2} (x_s - \frac{\alpha_{t|s} \sigma_s^2 x_t}{\sigma_t^2})^T (y_i - \bar{y}(x_t, t)) \\
    & ~~~~+ \frac{\alpha_s^2 \sigma_{s|t}^2}{2\sigma_s^4} (\bar{y}^2(x_t,t) - y_i^2) ) \\
    & = 1 + \frac{\alpha_s}{\sigma_s^2} (x_s - \frac{\alpha_{t|s} \sigma_s^2 x_t}{\sigma_t^2})^T (y_i - \bar{y}(x_t, t)) \\
    & ~~~~+ \frac{\alpha_s^2 \sigma_{s|t}^2}{2\sigma_s^4} (\bar{y}^2(x_t,t) - y_i^2) \\
    & ~~~~+ o(\frac{\alpha_s}{\sigma_s^2} 2 \sqrt{\sigma_{s|t}}M +  \frac{\alpha_s^2 \sigma_{s|t}^2}{2\sigma_s^4}2M^2) \\
    & = 1 + \frac{\alpha_s}{\sigma_s^2} (x_s - \frac{\alpha_{t|s} \sigma_s^2 x_t}{\sigma_t^2})^T (y_i - \bar{y}(x_t, t)) + o(\sqrt{\sigma_{s|t}}).
\end{aligned}
\end{equation}
Thus
\begin{equation}
\begin{aligned}
    &|p(x_s | x_t) - \tilde{p}(x_s | x_t)| \\
    &=\tilde{p}(x_s | x_t) | \frac{p(x_s | x_t)}{\tilde{p}(x_s | x_t)} - 1| \\
    &=\tilde{p}(x_s | x_t) | \sum_i w_i(x_t, t) \exp(\frac{C_i}{2 \sigma_{s|t}^2}) - 1| \\
    &=\tilde{p}(x_s | x_t) |\frac{\alpha_s}{\sigma_s^2} (x_s - \frac{\alpha_{t|s} \sigma_s^2 x_t}{\sigma_t^2})^T (\sum_i w_i(x_t, t) y_i - \bar{y}(x_t, t)) \\
    &~~~~+ o(\sqrt{\sigma_{s|t}})| \\
    &=\tilde{p}(x_s | x_t)(o(\sqrt{\sigma_{s|t}})),
\end{aligned}
\end{equation}
which means $\exists \tau_3 \in (s,1], C_1 >0$, such that $|p(x_s | x_t) - \tilde{p}(x_s | x_t)| < \tilde{p}(x_s | x_t) * C_1 \sqrt{\sigma_{s|t}}$, $\forall t \in (s,\tau_3]$

Let $\tau = \min(\tau_1, \tau_2, \tau_3)$. As a result
\begin{equation}\label{eq:prop1_bound}
\begin{aligned}
    &\int_{\mathbb{R}^{d}} |p(x_s| x_t) - \tilde{p}(x_s| x_t)| d x_s \\
    &= \int_{A(x_t, \bar{y}(x_t, t), 1)} |p(x_s| x_t) - \tilde{p}(x_s| x_t)| d x_s \\
    &~~~~+ \int_{A^c(x_t, \bar{y}(x_t, t), 1)} |p(x_s| x_t) - \tilde{p}(x_s| x_t)| d x_s \\
    &\leq 5\sigma_{s|t} + \int_{A^c(x_t, \bar{y}(x_t, t), 1)} \tilde{p}(x_s | x_t) * C_1 \sqrt{\sigma_{s|t}} ds \\
    & \leq 5\sigma_{s|t} + C_1 \sqrt{\sigma_{s|t}} < (5 + C_1)\sqrt{\sigma_{s|t}}
\end{aligned}
\end{equation}
Let $C = 5 + C_1$, this is the constant required in this proposition.  \hfill $\square$

\begin{Proposition_proof}
    $\exists \nu \in (0, 1)$ and $C > 0$, such that $\forall \nu \leq s < t \leq 1$,
$\int_{\mathbb{R}^{d}} |p(x_s| x_t) - \tilde{p}(x_s| x_t)| d x_s < C \sqrt{\alpha_s}$.
\end{Proposition_proof}

 \paragraph{Proof:}
 Let $A(x_t, y, c)$ denote the set $\{x_s : |x_s - \frac{\alpha_{t|s} \sigma_s^2 x_t}{\sigma_t^2} - \frac{\alpha_s \sigma_{s|t}^2 y}{\sigma_s^2}| > c \frac{1}{\sqrt{\alpha_s}}\}$.
\begin{equation}
\begin{aligned}
    &\tilde{P}(A(x_t, \bar{y}(x_t, t), 1) | x_t) \\
    &= \tilde{P}(|x_s - \frac{\alpha_{t|s} \sigma_s^2 x_t}{\sigma_t^2} - \frac{\alpha_s \sigma_{s|t}^2 \bar{y}(x_t, t)}{\sigma_s^2}| > \frac{1}{\sqrt{\alpha_s}} | x_t) \\
    &\leq \sigma_{s|t}^2 \alpha_s \leq \alpha_s.
\end{aligned}
\end{equation}
The last inequality is from the Chebyshev inequality.
Let $\nu_2$ be the value in Lemma 4(2), then $\forall \nu_2 \leq s < t \leq t$, $x_s \in A(x_t, \bar{y}(x_t, t), 1)$ indicates $x_s \in A(x_t, y_i, \frac{1}{2}), 1 \leq i \leq N$,
then
\begin{equation}
\begin{aligned}
    &P(A(x_t, \bar{y}(x_t, t), 1) | x_t, y_i) \\
    &\leq P(A(x_t, y_i, \frac{1}{2}) | x_t, y_i) \\
    &= P(|x_s - \frac{\alpha_{t|s} \sigma_s^2 x_t}{\sigma_t^2} - \frac{\alpha_s \sigma_{s|t}^2 y_i}{\sigma_s^2}| > \frac{1}{2 \sqrt{\alpha_s}} | x_t, y_i) \\
    &\leq \sigma_{s|t}^2 4\alpha_s \leq 4\alpha_s.
\end{aligned}
\end{equation}
As a result
\begin{equation}
\begin{aligned}
    &\int_{A(x_t, \bar{y}(x_t, t), 1)} |p(x_s| x_t) - \tilde{p}(x_s| x_t)| d x_s \\
    &\leq \sum_i \int_{A(x_t, \bar{y}(x_t, t), 1)} w_i(x_t, t)|p(x_s| x_t, y_i) - \tilde{p}(x_s| x_t)| d x_s \\
    &\leq \sum_i \int_{A(x_t, \bar{y}(x_t, t), 1)} w_i(x_t, t)(p(x_s| x_t, y_i) + \tilde{p}(x_s| x_t)) d x_s \\
    &\leq \sum_i  w_i(x_t, t) 5 \alpha_s = 5\alpha_s.
\end{aligned}
\end{equation}

On the other hand, if $x_s \in A^c (x_t, \bar{y}(x_t, t), 1)$. 
Let $\nu_1$ be the value in Lemma 4(1), $\forall \nu_2 \leq s < t \leq 1$,
then $x_s \in A^c (x_t, \bar{y}(x_t, t), 1)$ indicates $x_s \in A^c (x_t, 0, 2)$.
Denoting
\begin{equation}
\begin{aligned}
    C_i &= |x_s - \frac{\alpha_{t|s} \sigma_s^2 x_t}{\sigma_t^2} - \frac{\alpha_s \sigma_{s|t}^2 \bar{y}(x_t, t)}{\sigma_s^2}| ^ 2 \\
    & ~~~~- |x_s - \frac{\alpha_{t|s} \sigma_s^2 x_t}{\sigma_t^2} - \frac{\alpha_s \sigma_{s|t}^2 y_i}{\sigma_s^2}|^2,
\end{aligned}
\end{equation}
we have
\begin{equation}
\begin{aligned}
    |\frac{C_i}{2 \sigma_{s|t}^2}| &= |\frac{1}{2 \sigma_{s|t}^2} (|x_s - \frac{\alpha_{t|s} \sigma_s^2 x_t}{\sigma_t^2} - \frac{\alpha_s \sigma_{s|t}^2 \bar{y}(x_t, t)}{\sigma_s^2}| ^ 2 \\
    & ~~~~- |x_s - \frac{\alpha_{t|s} \sigma_s^2 x_t}{\sigma_t^2} - \frac{\alpha_s \sigma_{s|t}^2 y_i}{\sigma_s^2}|^2)| \\
    & = |\frac{\alpha_s }{\sigma_s^2} (x_s - \frac{\alpha_{t|s} \sigma_s^2 x_t}{\sigma_t^2})^T (y_i - \bar{y}(x_t, t)) \\
    & ~~~~+ \frac{\alpha_s^2 \sigma_{s|t}^2}{2\sigma_s^4} (\bar{y}^2(x_t,t) - y_i^2)| \\
    & \leq \frac{\sqrt{\alpha_s}}{\sigma_s^2} 2M +  \frac{\alpha_s^2 \sigma_{s|t}^2}{2\sigma_s^4}2M^2,
\end{aligned}
\end{equation}
then
\begin{equation}
\begin{aligned}
    \exp&(\frac{C_i}{2 \sigma_{s|t}^2}) = \exp( \frac{\alpha_s }{\sigma_s^2} (x_s - \frac{\alpha_{t|s} \sigma_s^2 x_t}{\sigma_t^2})^T (y_i - \bar{y}(x_t, t)) \\
    & ~~~~+ \frac{\alpha_s^2 \sigma_{s|t}^2}{2\sigma_s^4} (\bar{y}^2(x_t,t) - y_i^2) ) \\
    & = 1 + \frac{\alpha_s}{\sigma_s^2} (x_s - \frac{\alpha_{t|s} \sigma_s^2 x_t}{\sigma_t^2})^T (y_i - \bar{y}(x_t, t)) \\
    & ~~~~+ \frac{\alpha_s^2 \sigma_{s|t}^2}{2\sigma_s^4} (\bar{y}^2(x_t,t) - y_i^2) \\
    & ~~~~+ o(\frac{\sqrt{\alpha_s}}{\sigma_s^2} 2M + \frac{\alpha_s^2 \sigma_{s|t}^2}{2\sigma_s^4}2M^2) \\
    & = 1 + \frac{\alpha_s}{\sigma_s^2} (x_s - \frac{\alpha_{t|s} \sigma_s^2 x_t}{\sigma_t^2})^T (y_i - \bar{y}(x_t, t)) + o(\sqrt{\alpha_s}).
\end{aligned}
\end{equation}
Thus
\begin{equation}
\begin{aligned}
    &|p(x_s | x_t) - \tilde{p}(x_s | x_t)| \\
    &=\tilde{p}(x_s | x_t) | \frac{p(x_s | x_t)}{\tilde{p}(x_s | x_t)} - 1| \\
    &=\tilde{p}(x_s | x_t) | \sum_i w_i(x_t, t) \exp(\frac{C_i}{2 \sigma_{s|t}^2}) - 1| \\
    &=\tilde{p}(x_s | x_t) |\frac{\alpha_s}{\sigma_s^2} (x_s - \frac{\alpha_{t|s} \sigma_s^2 x_t}{\sigma_t^2})^T (\sum_i w_i(x_t, t) y_i  \\
    &~~~~- \bar{y}(x_t, t)) + o(\sqrt{\alpha_s})| \\
    &=\tilde{p}(x_s | x_t)(o(\sqrt{\alpha_s})),
\end{aligned}
\end{equation}
which means $\exists \nu_3 \in (s,1], C_1 > 0$, such that $|p(x_s | x_t) - \tilde{p}(x_s | x_t)| < \tilde{p}(x_s | x_t) C_1 \sqrt{\alpha_s}$, $\forall t \in (s,\nu_3]$.

Let $\nu = \min(\nu_1, \nu_2, \nu_3)$. As a result
\begin{equation}
\begin{aligned}
    &\int_{\mathbb{R}^{d}} |p(x_s| x_t) - \tilde{p}(x_s| x_t)| d x_s \\
    &= \int_{A(x_t, \bar{y}(x_t, t), 1)} |p(x_s| x_t) - \tilde{p}(x_s| x_t)| d x_s \\
    &~~~~+ \int_{A^c(x_t, \bar{y}(x_t, t), 1)} |p(x_s| x_t) - \tilde{p}(x_s| x_t)| d x_s \\
    &\leq 5\alpha_s + \int_{A^c(x_t, \bar{y}(x_t, t), 1)} \tilde{p}(x_s | x_t) * C_1 \sqrt{\alpha_s} ds \\
    &\leq 5\alpha_s + C_1 \sqrt{\alpha_s} < (5 + C_1)\sqrt{\alpha_s}
\end{aligned}
\end{equation}
Let $C = 5 + C_1$, this is the constant required in this proposition.  \hfill $\square$

\textit{Remark.} Observing that the two parts of the bound in Eq. \ref{eq:prop1_bound} is $\sigma_{s|t}$ and $\sqrt{\sigma_{s|t}}$, which have different order. We can balance the two orders by setting the set $A(x_t, y, c)$ to be $\{ x_s : |x_s - \frac{\alpha_{t|s} \sigma_s^2 x_t}{\sigma_t^2} - \frac{\alpha_s \sigma_{s|t}^2 y}{\sigma_s^2}| > c (\sigma_{s|t})^\frac{2}{3}\}$. With the same proof process, we can get a bound $ (\sigma_{s|t})^\frac{2}{3}$, which is better than $\sqrt{\sigma_{s|t}}$. Similarly, we can also get a better bound for Proposition 2.

\begin{Proposition_proof}
Setting $x_{1-\epsilon} \sim \mathcal{N}(0, I)$ is equivalent to sampling the value from standard Gaussian as $\bar{y}(x_{1}, 1)$ at $t=1$.
\end{Proposition_proof}

\paragraph{Proof:} We will demonstrate this proposition in two scenarios, DDPM and DDIM.

\noindent\textbf{(1) Sampling by DDPM:}

The sampling step for DDPM is
\begin{equation} \label{eq:prop3_ddpm}
    x_s = \frac{\alpha_{t|s} \sigma_s^2}{\sigma_t^2} x_t + \frac{\alpha_s \sigma_{t|s}^2}{\sigma_t^2} \bar{y}(x_t, t) + \sigma_{s|t} z_t.
\end{equation}
Applying the conditions that $\alpha_1 = 0$ and $\sigma_1 = 1$ to this equation, we obtain
\begin{equation} \label{eq:prop3_ddpm1}
    x_{1 - \epsilon} = \alpha_{1 - \epsilon} \bar{y}(x_1, 1) + \sigma_{1 - \epsilon} z_1,
\end{equation}
where $z_1$ follows a standard Gaussian distribution. Assuming that $x_{1-\epsilon} \sim \mathcal{N}(0, I)$ and is independent of $z_1$, it leads us to conclude that
\begin{equation}
    \bar{y}(x_1, 1) = \frac{x_{1 - \epsilon} - \sigma_{1 - \epsilon} z_1}{\alpha_{1-\epsilon}},
\end{equation}
also follows a Gaussian distribution. Given that $\alpha_{1-\epsilon}^2 + \sigma_{1 - \epsilon}^2 = 1$, we can deduce the following quantities
\begin{equation}
\begin{aligned}
    E(\bar{y}(x_1, 1)) = \frac{0 - \sigma_{1 - \epsilon} 0}{\alpha_{1-\epsilon}} = 0, \\
    Var(\bar{y}(x_1, 1)) = \frac{I - \sigma_{1 - \epsilon}^2 I}{\alpha_{1-\epsilon}^2} = I.
\end{aligned}
\end{equation}
This implies that $\bar{y}(x_1, 1)$ follows a standard Gaussian distribution.

\noindent\textbf{(2) Sampling by DDIM:}

The sampling step for DDPM from $1$ to $1-\epsilon$ is
\begin{equation}
    x_{1-\epsilon} = \alpha_{1-\epsilon} \bar{y}(x_1, 1) + \sigma_{1-\epsilon} x_1.
\end{equation}
From this, we can deduce that
\begin{equation}
    \bar{y}(x_1, 1) = \frac{x_{1-\epsilon} - \sigma_{1-\epsilon} x_1}{\alpha_{1-\epsilon}},
\end{equation}
is normally distributed.
Assuming that $x_{1-\epsilon} \sim \mathcal{N}(0, I)$ and is independent of $x_1$, we can calculate the following quantities
\begin{equation}
\begin{aligned}
    E(\bar{y}(x_1, 1)) = \frac{0 - \sigma_{1 - \epsilon} 0}{\alpha_{1-\epsilon}} = 0, \\
    Var(\bar{y}(x_1, 1)) = \frac{I - \sigma_{1 - \epsilon}^2 I}{\alpha_{1-\epsilon}^2} = I.
\end{aligned}
\end{equation}
Therefore, $\bar{y}(x_1, 1)$ is confirmed to be a standard Gaussian distribution.

\hfill $\square$

\setlength{\tabcolsep}{1.0mm}{
\begin{table}[t]
\centering
\renewcommand{\arraystretch}{1.1}
\caption{Comparison of average brightness of 100 generated images between SD-2.0 and our SingDiffusion under different prompt conditions. For 'white' prompts, higher average brightness is preferable, while for 'black' prompts, lower average brightness is better.}
\label{tab:gray_app}
\begin{tabular}{c|c|c|c|c}
\Xhline{1pt}
        \small Model & \begin{tabular}[c]{@{}c@{}} \scriptsize"Solid \textbf{white} \vspace{-8pt} \\ \scriptsize background"\end{tabular} & \begin{tabular}[c]{@{}c@{}}\scriptsize"Solid \textbf{black} \vspace{-8pt} \\ \scriptsize background"\end{tabular} & \begin{tabular}[c]{@{}c@{}c@{}}\scriptsize"Monochrome \scriptsize \vspace{-8pt} \\ \scriptsize line-art logo on a \vspace{-8pt} \\ \scriptsize \textbf{white} background"\end{tabular} & \begin{tabular}[c]{@{}c@{}c@{}}\scriptsize"Monochrome \scriptsize \vspace{-8pt} \\ \scriptsize line-art logo on a \vspace{-8pt} \\ \scriptsize \textbf{black} background"\end{tabular}  \\ \Xhline{1pt}
\small SD-2.0 &        137.64                  &           84.94               &                                                          130.68                                   &      105.07                                                                                       \\
\small + Ours  &         \textbf{231.26}                 &            \textbf{1.54}              &    \textbf{238.67}                                                                                         &      \textbf{9.22}                                                                                       \\ \Xhline{1pt}
\end{tabular}
\end{table}
}

\setlength{\tabcolsep}{2.2mm}{
\begin{table}[!t]
\centering
\renewcommand{\arraystretch}{1.1}
\caption{Comparison of SD-2.0 model and SingDiffusion on FID score and CLIP score without classifier guidance.}
\begin{tabular}{c|cc}
\Xhline{1pt} Model & FID $\downarrow$ & CLIP $\uparrow$   \\ \Xhline{1pt}
Original & 36.42 & 27.46   \\
+ SingDiffusion & \textbf{20.13} & \textbf{28.43}   \\ \Xhline{1pt}
\end{tabular}
\label{tab:FID_app}
\end{table}
}

\begin{figure}[t]
\centering
\includegraphics[width=0.48\textwidth]{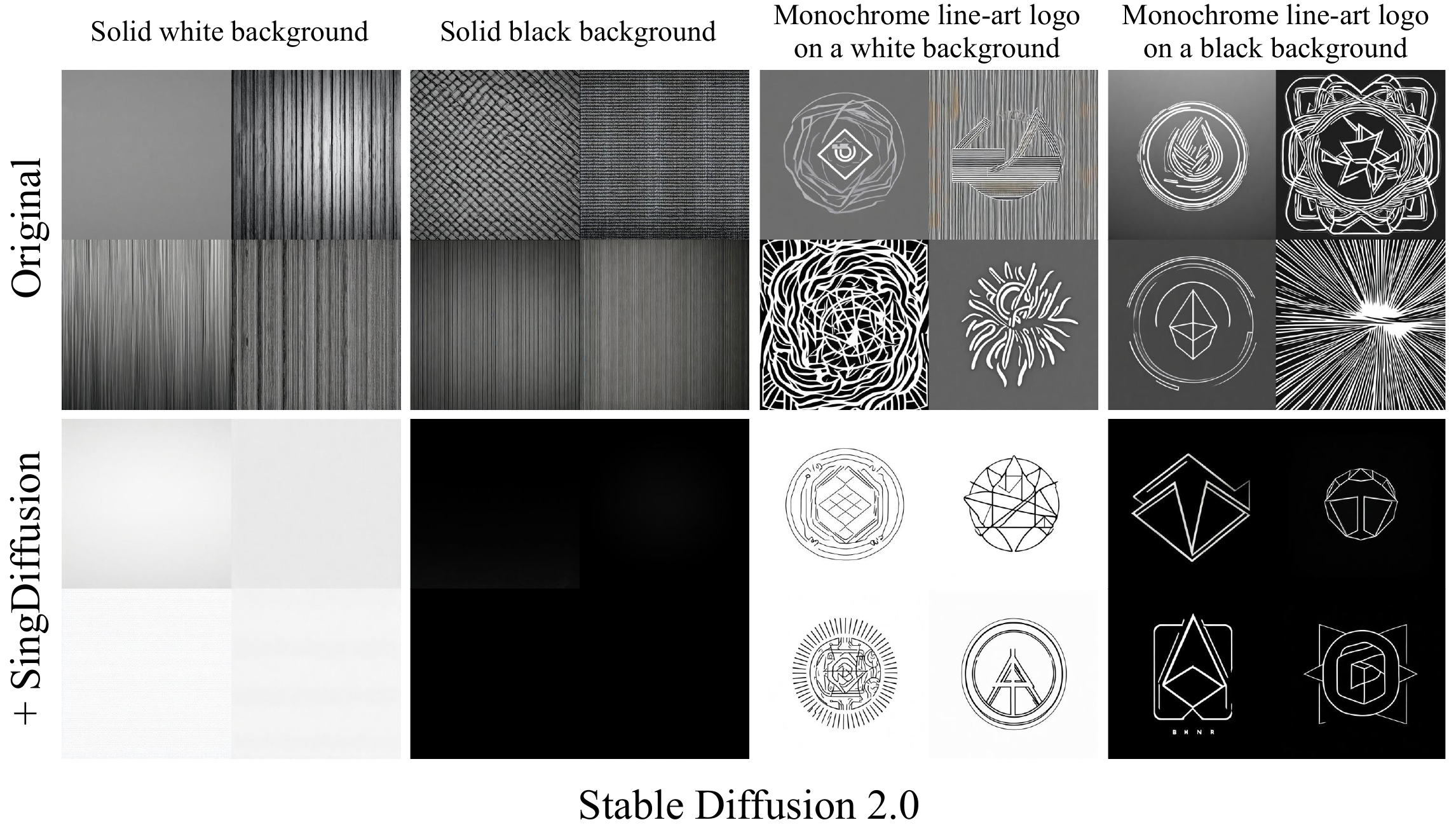} 
\caption{Comparison of SD-2.0 and our SingDiffusion on average brightness issue.}
\label{fig:black_app}
\end{figure}

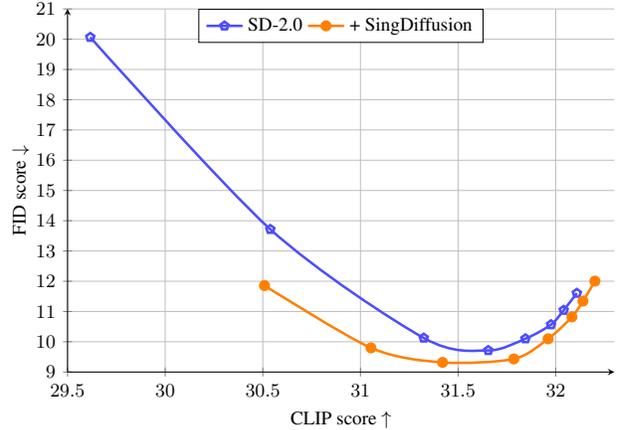
\begin{figure}[!t]
\centering
\center
\begin{tikzpicture}[font=\small,scale=0.8]
         \begin{axis}[
            ymin=9, ymax=21,
            ytick={9,10,...,21},
            xmin=29.5, xmax=32.3,
            ylabel=y, 
            ylabel=FID score $\downarrow$, y label style={at={(0.05,0.5)}},
               axis y line=left,
               axis x line=bottom, xlabel=CLIP score $\uparrow$, 
            width=4.2in,
            height=3.0in,
            grid=major,
            legend style={at={(0.5,1.0)},anchor=north}, legend columns=2]
            \addlegendentry{SD-2.0}
             \addplot[smooth, mark=pentagon,blue!70!, mark size =2pt, very thick] plot coordinates {
             (29.6165027618408,20.0642996186014)
             (30.5385189056396,13.7154312152213)
             (31.3239974975585,10.1255497331941)
             (31.6545448303222,9.71380372553034)
             (31.8447341918945,10.101257103721)
             (31.9758319854736,10.567111632059)
             (32.0405807495117,11.0427920538184)
             (32.1079673767089,11.6060571149988)
             };
             \addplot[smooth, mark=*,orange, mark size =2pt, very thick] plot coordinates {
             (30.5086402893066,11.8573212833379)
             (31.0539665222167,9.79814273970237)
             (31.420274734497,9.31826633486105)
             (31.7853736877441,9.43308200054787)
             (31.9607982635498,10.0966690595709)
             (32.0828475952148,10.8211334642969)
             (32.1391372680664,11.3442816161577)
             (32.2005653381347,12.0048175095428)
             };
             \addlegendentry{+ SingDiffusion}
     \end{axis}
     \end{tikzpicture}
\caption{Comparison of Pareto curves between SD-2.0 and our SingDiffusion on 30k COCO images, across various guidance scales in [1.5, 2, 3, 4, 5, 6, 7, 8].}
\label{fig:curve_app}
\end{figure}

\section{Application on SD-2.0}
Stable diffusion 2.0 model (SD-2.0) samples images using $v$-prediction, which differs from SD-2.0-base that utilizes $\epsilon$-prediction. To validate the applicability of our SingDiffusion method to the $v$-prediction approach, we conducted identical experiments on SD-2.0 as with SD-2.0-base. Firstly, we present the outcomes of our method in addressing the average brightness issue on SD-2.0, as depicted in Figure \ref{fig:black_app} and Table \ref{tab:gray_app}. From the visualizations, it is evident that our method significantly alleviates SD-2.0's challenge in generating both bright and dark images. Furthermore, we compare the generation capabilities of SD-2.0 on the COCO dataset before and after applying our method, as detailed in Table \ref{tab:FID_app}. It is noticeable that our method substantially reduces the FID score, resulting in more realistic outcomes. Lastly, we compared the CLIP vs. FID Pareto curves of the two methods across various guidance scales. As illustrated in Figure \ref{fig:curve_app}, our method achieves lower FID scores for the same CLIP score.


\end{document}